\documentclass[a4paper, 10pt, conference]{ieeeconf}      

\IEEEoverridecommandlockouts                              
\usepackage{graphicx} 
\usepackage{amsmath} 
\usepackage{amssymb}  
\usepackage{multirow}
\usepackage{booktabs}
\usepackage{threeparttable}
\usepackage{algorithm}
\usepackage{algorithmic}
\usepackage{pifont}
\usepackage{tabularx}
\let\labelindent\relax
\usepackage{enumitem}
\setlist[itemize]{leftmargin=*, topsep=2pt, itemsep=1pt, parsep=0pt}
\newcommand{\xmark}{\ding{55}}

\title{\LARGE \bf
ReMiX-MAE: Learning Missing-Channel Cross-Modal\\
Representations from RGB-Only Clinical Facial Videos \\
for Sympathetic-Mediated Pain Assessment
}

\author{\parbox{16cm}{\centering
    {\large Nan Bi$^1$, Taoyue Wang$^1$, Lijun Yin$^1$ and Vandana Sharma$^2$}\\
    {\normalsize
    $^1$ School of Computing, Binghamton University, Binghamton, NY, USA\\
    $^2$ Department of Anesthesiology, SUNY Upstate Medical University, Syracuse, NY, USA}}
    \thanks{This work was supported by Watson College-Upstate Medical Pilot Research Grant Program and NSF(CNS-1629898)}
}
\begin{document}

\maketitle
\thispagestyle{empty}
\pagestyle{empty}

\begin{abstract}
Automated pain assessment in real clinics is limited by scarce clinically grounded facial video data with weak labels (often sequence-level self-report) and by the fact that pain cues can be subtle or near-neutral in RGB, while thermal/depth signals are informative yet impractical to deploy routinely. To address these challenges, we propose ReMiX-MAE (Reconstructing Missing-Channel Cross-Modal Masked Autoencoder), a self-supervised multimodal masked pretraining framework that learns transferable facial representations from synchronized RGB, thermal, and depth videos and explicitly trains robustness to missing modalities, enabling RGB-only deployment. To fill the gap of clinically grounded facial pain data with video-level self-report and longitudinal treatment trajectories, we collect the Sympathetic-Mediated Pain (SMP) dataset with paired pre/post recordings across multiple visits.
 Under RGB-only deployment, we evaluate ReMiX-MAE using both direct feature extraction and pseudo-multimodal features decoded from RGB. ReMiX-MAE consistently outperforms an RGB-only masked autoencoder baseline on SMP, with pseudo-multimodal features providing additional gains in the challenging five-class setting. Across external datasets, ReMiX-MAE further shows more robust and label-efficient transfer than RGB-only baselines, highlighting its advantage in data-limited clinical settings.
\end{abstract}

\section{INTRODUCTION}

Pain severity is most meaningfully grounded in patient self-report, commonly measured by the Visual Analog Scale (VAS) or the Numeric Rating Scale (NRS), which remains central to clinical decision-making. 
 However, self-report is not always available, timely, or practical in routine care, and pain trajectories can fluctuate within and across clinical encounters~\cite{dobscha2015short,schneider2012individual}. Pain assessment is also prone to systematic biases in both self- and other-report, motivating more objective estimates from routine recordings~\cite{craig1984observer,williams2002facial,trawalter2012racial,mende2019perceptual}. These limitations have motivated growing interest in automated pain assessment methods that estimate pain-related states from observable physiological and behavioral cues. Rather than replacing patient-reported outcomes, such systems aim to complement them by providing continuous, scalable, and reproducible monitoring signals that can assist clinicians when self-report is missing, delayed, or difficult to obtain. Among available signals, facial expressions are particularly compelling because they are contact-free, easily captured, and closely linked to affective and autonomic states.

Analyzing facial behavior from RGB images or videos provides a contact-free and readily deployable solution for real-world clinical use. In this context, the Facial Action Coding System (FACS)~\cite{au} has enabled the widely adopted Prkachin and Solomon Pain Intensity (PSPI) score~\cite{prkachin2008pspi}, which offers frame-level estimates of pain-related facial action. Beyond facial video, a line of work has explored additional physiological measurements—such as galvanic skin response, photoplethysmography, electrodermal activity, and respiratory effort~\cite{walter2013biovid,ai4pain,emopain} —to improve pain inference. While informative, such contact-based measurements typically require specialized hardware and skin contact, limiting scalability and routine clinical deployment.

Recent clinical studies~\cite{thermalvision, thermalvision2, thermalvision3} suggest a more practical, camera-based alternative: thermal and depth imaging. These modalities can capture pain-related changes in cutaneous perfusion and temperature patterns as well as subtle three-dimensional facial micro-movements, providing complementary information beyond RGB. However, an important hurdle remains. Whereas RGB cameras are ubiquitous, dedicated thermal and depth sensors are frequently unavailable in routine clinical workflows, creating a gap between multimodal research settings and RGB-only deployment in practice.

To bridge this gap, we leverage multimodal data (RGB, thermal, and depth) only during representation learning, while maintaining RGB-only inference to preserve practicality at deployment. We propose ReMiX-MAE (Reconstructing Missing-Channel Cross-Modal Masked Autoencoder), a self-supervised framework that learns transferable facial representations by reconstructing masked regions while explicitly simulating missing modalities via dynamic channel routing. By training the model to operate under modality-missing conditions, ReMiX-MAE leverages thermal and depth as complementary training-time signals to learn representations that remain effective under \emph{RGB-only} deployment. The pretrained encoder is subsequently used as a frozen feature extractor paired with lightweight downstream heads, enabling data-efficient learning in small clinical datasets and facilitating transfer across heterogeneous cohorts.

Progress in automated pain assessment also depends on the availability of clinically relevant datasets, yet public resources remain limited. The UNBC-McMaster Shoulder Pain dataset (UNBC)~\cite{ShoulderPain} is widely used in the literature, containing video sequences from a small number of patients with corresponding annotations. Much prior work has focused on frame-level PSPI labels rather than video-level self-report pain score (VAS), in part because the number of annotated sequences is limited. Importantly, based on the study in~\cite{prkachin2008pspi}, PSPI is statistically related to VAS but only moderately correlated (0.36-0.37), suggesting that estimating VAS may better align with clinically meaningful, patient-reported pain severity. In this work, we use UNBC’s provided sequence-level self-reported VAS scores as ground truth (not PSPI-derived expression intensity).

In contrast, datasets collected under tightly controlled laboratory conditions, where acute experimental pain is induced in healthy participants via heat or pressure, are typically easier to acquire at scale and therefore substantially larger. BioVid~\cite{walter2013biovid} is a representative example, providing thousands of videos with well-defined pain stimuli at multiple intensities and accompanying physiological recordings. However, prior work~\cite{prajod2024faces,prajod2022using} has shown that models trained exclusively on experimental datasets often exhibit a marked performance drop when transferred to clinical data, indicating a substantial domain gap and suggesting that salient facial cues may differ across experimental and clinical settings.

Motivated by this gap, we collect a clinical pain video dataset focusing on patients experiencing chronic SMP. Compared with UNBC, the observed distribution of video-level self-report pain score in our dataset is less skewed, and although we include fewer sequences, our recordings are longer and therefore yield a larger total number of frames. In contrast to experimental datasets such as BioVid, our cohort is clinically grounded. To our knowledge, this is the first clinical facial pain video dataset that provides within-subject longitudinal recordings before and after treatment, enabling analysis of pain-related facial dynamics across therapeutic change and clinically aligned evaluation of automated pain assessment models.

Our work makes the following primary contributions:
\begin{enumerate}
     \item \textbf{Method:}
    We propose ReMiX-MAE, a self-supervised cross-modal masked autoencoder that distills complementary thermal/depth cues into a channel-agnostic encoder via explicit missing-modality simulation, enabling RGB-only inference. We validate its effectiveness through targeted ablations and comparisons to strong supervised and self-supervised baselines, as well as transfer experiments across experimental and clinical datasets.

    \item \textbf{Dataset:}
    We collect a clinical chronic pain video dataset with video-level self-report pain score annotations and, to our knowledge, the first within-subject pre-/post-treatment facial recordings in a clinical pain corpus. The dataset supports self-report pain score evaluation and longitudinal analysis of whether automated predictions track therapeutic change.
\end{enumerate}

\section{Related Work}
\label{sec:related_work}

\subsection{Facial Pain Assessment: Datasets and Methods}
Automatic pain assessment from facial behavior has been studied extensively, motivated by the need to support vulnerable or non-communicative patients. Prior surveys highlight persistent challenges for clinical deployment, including ambiguities in ground-truth definition, the scarcity of clinically validated datasets, and the relative lack of focus on chronic pain compared to acute or stimulus-driven settings~\cite{werner_pain_survey}.

Facial videos captured in clinical environments are highly variable in pose, illumination, and background, and are typically annotated using patient self-report or clinical pain scales. For example, Liu \emph{et al.}~\cite{liu2018clinical} collected an emergency-room dataset with RGB and Kinect v2 videos, paired with sequence-level self-reported pain scores and biochemical markers (COX-2 and iNOS), to link facial behavior to clinical pain context.

A substantial body of work evaluates pain recognition methods on the UNBC-McMaster Shoulder Pain Expression Archive~\cite{ShoulderPain}, which provides frame-level FACS-based~\cite{au} annotations and PSPI scores~\cite{prkachin2008pspi}, as well as sequence-level self-reported pain score. Most studies treat PSPI as the primary supervision signal and model pain at the frame level, with representative approaches spanning handcrafted descriptors with regression~\cite{kaltwang2012continuous,werner2014automatic}, temporal modeling with recurrent networks~\cite{zhou2016recurrent}, and feature fusion strategies that combine heterogeneous representations~\cite{yang2018lowhigh,deeppain,xu2021deep,ting2021distance,huang2022hybnet}. Beyond appearance cues, non-contact physiological correlates have also been explored: Yang \emph{et al.}~\cite{yang2021rstan} predicted remote photoplethysmography (rPPG) from facial videos and leveraged it as an auxiliary task, reporting improved robustness on UNBC and BioVid.

In contrast to clinically oriented datasets, experimental benchmarks such as BioVid~\cite{walter2013biovid} are collected under tightly controlled stimulus protocols, enabling larger-scale data collection and multimodal sensing. BioVid provides video-level labels corresponding to no-pain and four graded stimulus intensities (often used as a proxy for pain severity), together with synchronized physiological signals such as galvanic skin response (GSR), electrocardiography (ECG), and electromyography (EMG). Recent work reports strong performance on BioVid using modern video architectures and multimodal pipelines~\cite{gkikas2023full,huang2021spatio}. However, prior studies also show that models trained primarily on experimental datasets can exhibit marked performance drops when transferred to clinical data~\cite{prajod2024faces,prajod2022using}, highlighting the challenge of generalization across recording conditions, label definitions, and pain settings.

\subsection{Self-Supervised and Multimodal Video Representation Learning}
Self-supervised masked modeling has become a strong paradigm for learning transferable visual representations without manual labels. VideoMAE-style masked video autoencoding learns spatio-temporal features by reconstructing masked tokens from partially observed inputs, providing a scalable pretraining approach for downstream recognition tasks~\cite{tong2022videomae}.

Frameworks such as MARLIN~\cite{cai2023marlin} extend masked modeling to facial representation learning, demonstrating that masked objectives can learn robust facial video encoders that transfer effectively to downstream affective tasks. Zhang \emph{et al.}~\cite{zhang2024multimodal} proposed Multimodal Channel-Mixing (MCM), a channel-dropping and reconstruction strategy for frame-level facial action unit detection. These developments motivate studying channel-missing reconstruction under spatio-temporal video modeling, particularly in settings where modality availability differs between training and deployment (e.g., RGB-only inference).

\section{SMP Video Dataset}
\label{sec:syracuse_dataset}

\subsection{Clinical Motivation and Dataset Overview}
Publicly available facial pain datasets with clinically grounded, video-level self-reported labels remain limited, and longitudinal recordings that capture treatment-related change are particularly rare. To address this gap, we collect the \textbf{SMP} dataset, a clinical facial video corpus of patients experiencing chronic SMP, designed to support (i) \emph{RGB-only} pain assessment under real outpatient conditions and (ii) \emph{within-subject} analysis of pain trajectories across treatment.

SMP presents a stringent testbed for vision-based pain modeling. Compared to experimentally induced pain, facial manifestations in SMP are often subtle, intermittent, and context-dependent, with many frames appearing near-neutral in RGB despite meaningful self-reported pain. Clinically, SMP is often associated with autonomic dysregulation and central sensitization, and its presentation can be heterogeneous across patients; as a result, pain may not consistently elicit overt facial actions.

\begin{figure}[t]
  \centering
  \includegraphics[width=\linewidth]{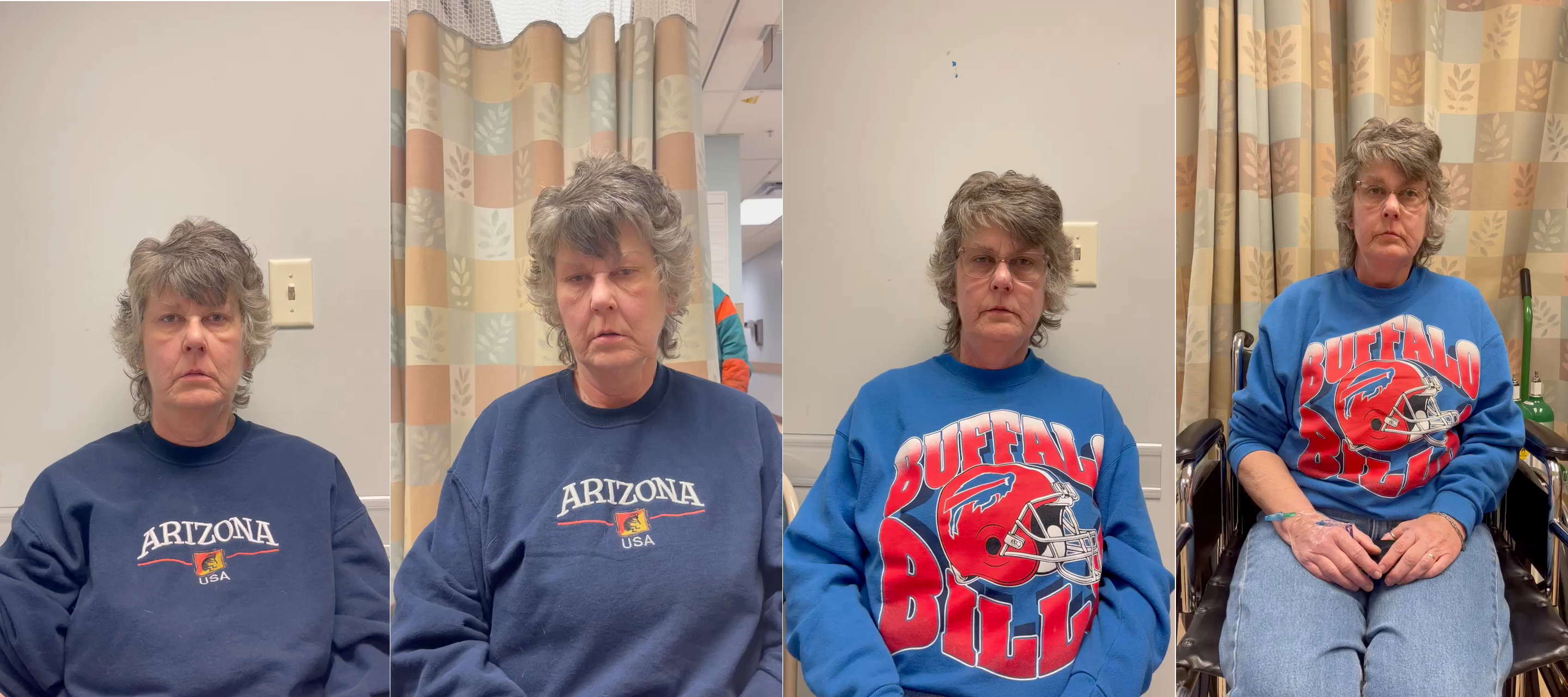}

\caption{SMP example for one subject (pre/post at initial and final visits). Pain scores: 9, 3, 2, 1.}
\label{fig:smp_example_prepost}

\end{figure}

Fig.~\ref{fig:smp_example_prepost} shows an illustrative within-subject example highlighting that large self-reported pain changes may correspond to only subtle visual differences in RGB. Overall, SMP contains recordings from \textbf{36} unique patients (24 female, 12 male), with up to four clips per patient across multiple clinically meaningful timepoints (not all participants have all four recordings due to clinical workflow and recording quality). In total, \textbf{72} video recordings were collected. Each clip is approximately one minute long, providing a longer observation window and a larger number of frames per sequence, which is beneficial when pain-related evidence is sparse or intermittent.

\subsection{Study Design, Ethics, and Longitudinal Protocol}
SMP was collected as part of a prospective observational study conducted at a university-affiliated pain management clinic, with institutional review board (IRB) approval and written informed consent from all participants.

Participants were adult patients scheduled for routine pain-relief injection procedures as part of standard clinical care (e.g., lumbar epidural steroid injections and sympathetic block injections). Importantly, the clinical workflow was not modified for data collection.

Each participant was recorded longitudinally at two clinically meaningful visits: an initial visit and a final follow-up visit in the care trajectory. At each visit, recordings were acquired \textbf{immediately before} the injection procedure and \textbf{shortly after} the procedure when applicable. This design yields up to four clips per patient (pre/post at the initial visit and pre/post at the final follow-up), enabling within-subject comparisons of both short-term treatment effects and longer-term changes across the treatment trajectory.

\subsection{Video Acquisition in Real-World Clinical Conditions}
All videos were captured in the outpatient clinic using a consumer-grade smartphone (iPhone) at 720p resolution and 30 fps without specialized imaging hardware, reflecting the practical constraints of routine clinical workflows. Recordings were collected under natural environmental conditions, resulting in unconstrained variations in pose, illumination, background, and patient appearance. No scripted expressions or elicited tasks were used; instead, SMP captures spontaneous facial behavior representative of real outpatient encounters and aligns with practical RGB-only deployment.


\subsection{Pain Labels and Dataset Statistics}
Immediately after each recording, patients reported their current pain intensity on NRS (0–10), where 0 indicates no pain and 10 indicates the worst imaginable pain. We use these sequence-level self-reports as the primary labels for supervised learning and evaluation on SMP. For clarity, UNBC reports pain using VAS, whereas SMP uses NRS; both are patient-reported intensity measures on a 0--10 scale, and we refer to them collectively as self-reported pain scores when the distinction is not essential.

\begin{figure}[t]
  \centering
  \includegraphics[width=\linewidth]{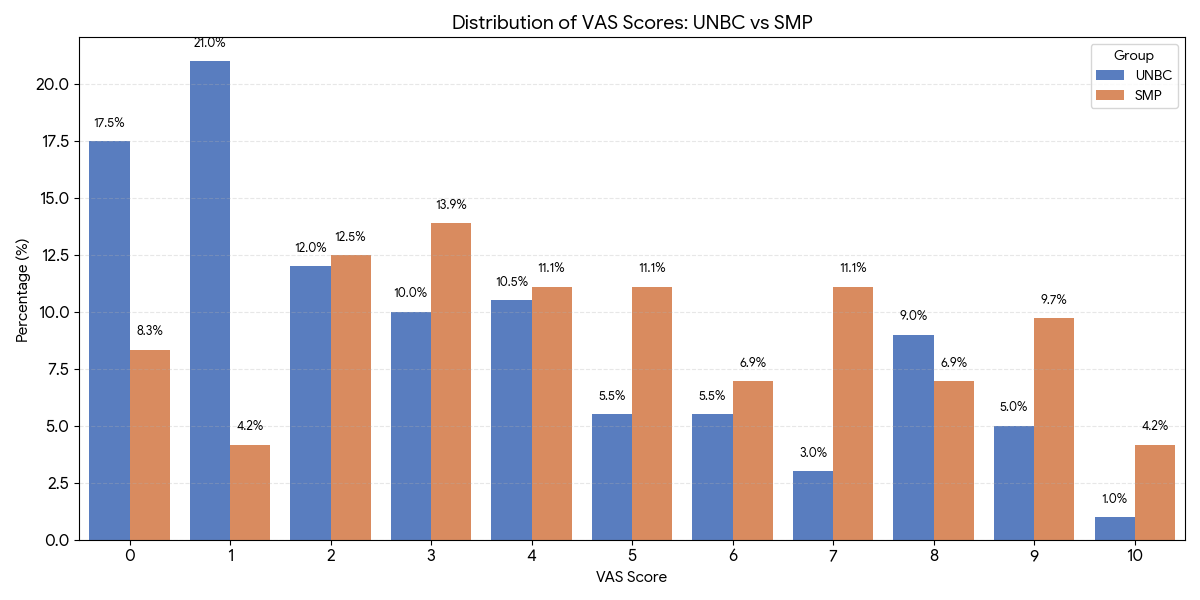}
  \caption{Distribution of self-reported pain intensity in SMP (orange) compared to UNBC (blue). SMP shows reduced label skew across the 0--10 scale, likely reflecting differences in cohort and pain construct (chronic clinical self-report vs controlled settings), as well as differences in recording protocol and supervision.}
  \label{fig:pain_distribution}
\end{figure}

Fig.~\ref{fig:pain_distribution} summarizes the distribution of self-reported pain levels in SMP and contrasts it with the UNBC distribution. Compared to UNBC, SMP shows \textbf{reduced label skew} across the 0--10 scale, which helps mitigate class imbalance in video-level learning and evaluation.

\subsection{Additional Clinical Measurements}
As part of the broader clinical study, blood samples were collected at selected visits in conjunction with injection procedures to support future analyses of biochemical markers associated with SMP. These measurements are not used for model training or evaluation in this work.

\section{The ReMiX-MAE Framework}
\label{sec:mma}

\subsection{Problem Setting and Overview}
Real-world clinical pain assessment is typically constrained to \emph{RGB-only} video, whereas thermal and depth streams can provide complementary physiological and geometric cues~\cite{levenson1990voluntary, kreibig2010autonomic} but are rarely available in routine workflows. This induces a practical \emph{training--deployment modality mismatch}: multimodal signals may be available for representation learning, yet deployment must often operate with RGB alone. ReMiX-MAE (Reconstructing Missing-Channel Cross-Modal Masked Autoencoder) addresses this mismatch with a self-supervised masked pretraining framework that (i) leverages synchronized RGB/thermal/depth videos when available, (ii) explicitly trains representations to be robust to missing modalities, and (iii) preserves compatibility with standard \emph{3-channel} video backbones and their pretrained initialization.

\noindent\textbf{Design choices.}
\textbf{(1) Patch-wise 5-to-3 channel routing} routes five atomic channels (R/G/B/depth/thermal) into three input slots at the patch level, enabling early fusion under a fixed 3-channel interface.
\textbf{(2) Modality-identity injection} adds a token-wise identity bias derived from which atomic modalities contributed to each token, disambiguating mixed-channel tokens.
\textbf{(3) Episode-based missing-modality simulation} samples episode types that define modality/channel missingness; the encoder input is gated \emph{only} by the MAE patch mask, while episode-defined declared-missing metadata specifies auxiliary channel-drop supervision without removing tokens from the encoder input.

\begin{figure*}[t]
  \centering
  \includegraphics[width=\linewidth]{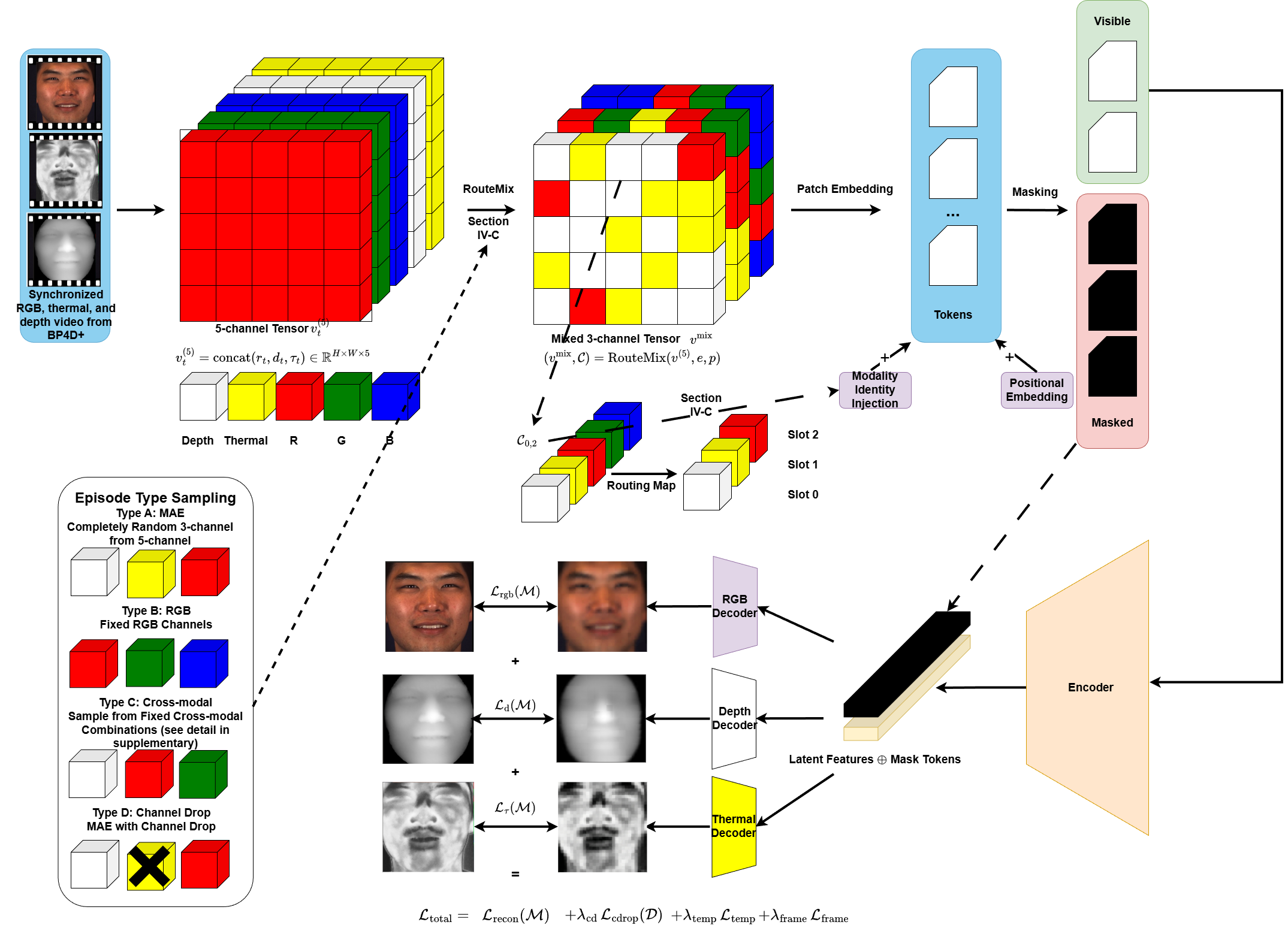}
  \caption{Overview of ReMiX-MAE (upstream pretraining). Synchronized RGB, thermal, and depth facial videos are routed from five atomic channels to a 3-channel mixed input at the patch level, tokenized into tubelets, and masked for reconstruction. A shared video encoder is trained with modality-specific decoders and episode-driven auxiliary supervision.}
  \label{fig:remix_framework}
\end{figure*}

Fig.~\ref{fig:remix_framework} illustrates the upstream pretraining pipeline. Algorithm~\ref{alg:routing_identity} defines \textsc{RouteMix} (routing and identity metadata), and Algorithm~\ref{alg:remix_pretrain} summarizes the episode-driven pretraining loop.

\subsection{Backbone Architecture}

Given a temporally aligned facial clip with $T$ frames, we denote the RGB, depth, and thermal clips as
$r\in\mathbb{R}^{T\times H\times W\times 3}$, $d\in\mathbb{R}^{T\times H\times W\times 1}$, and
$\tau\in\mathbb{R}^{T\times H\times W\times 1}$, respectively.
We concatenate them along the channel dimension to form a 5-channel clip
\[
v^{(5)}=\mathrm{concat}(r,d,\tau)\in\mathbb{R}^{T\times H\times W\times 5},
\]
corresponding to $\{R,G,B,\mathrm{D},\tau\}$.
A patch-wise 5-to-3 routing module (Sec.~\ref{subsec:routing}) selects three channels per patch and produces a mixed three-channel clip
$v^{\mathrm{mix}}\in\mathbb{R}^{T\times H\times W\times 3}$.

We tokenize $v^{\mathrm{mix}}$ into spatio-temporal tubelets using a 3D convolutional patch embedding with spatial patch size $p$
and temporal tubelet size $t$, yielding a sequence of $N=\frac{T}{t}\cdot\frac{H}{p}\cdot\frac{W}{p}$ tokens.
Let $\mathcal{P}=\{1,\dots,N\}$ denote the token index set. Following masked autoencoding, we sample a mask set $\mathcal{M}\subset\mathcal{P}$
and encode only the visible tokens $\mathcal{V}=\mathcal{P}\setminus\mathcal{M}$:
\[
z = f_\theta\!\left(\tilde{E}_{\mathcal{V}}\right),
\]
where $\tilde{E}$ denotes token embeddings augmented with positional encoding and modality-identity injection
(Section~\ref{subsec:routing}) to disambiguate heterogeneous mixed-channel tokens.

On top of the shared latent representation $z$, we attach modality-specific decoder heads for RGB, thermal, and depth, which map $z$
back to the corresponding modality spaces for reconstruction-style supervision. Denoting the decoder for modality $m$ by $g_{\phi_m}$,
we write:
\[
\hat{v}^{(\mathrm{rgb})} = g_{\phi_{\mathrm{rgb}}}(z),\qquad
(\hat{\mu}^{(m)}, \hat{s}^{(m)})=g_{\phi_m}(z)\ \ \text{for}\ \ m\in\{\tau,\mathrm{d}\},
\]
where for RGB, the decoder outputs patchified pixel intensities, while for thermal/depth, the decoder outputs Gaussian parameters $(\hat{\mu}^{(m)}, \hat{s}^{(m)})$ with
$\hat{s}=\log\sigma^2$ (log-variance), defining a per-token likelihood for $m\in\{\tau,\mathrm{d}\}$. The full pretraining objectives and auxiliary supervision
terms are described in Section~\ref{subsec:pretrain_objective}.

\subsection{RouteMix: Patch-wise Routing and Modality Identity Injection}
\label{subsec:routing}

\begin{algorithm}[t]
\caption{\textsc{RouteMix}: Patch-wise 5-to-3 routing with canonical slot identities}
\label{alg:routing_identity}
\begin{algorithmic}[1]
\REQUIRE 5-channel clip $v^{(5)}\in\mathbb{R}^{T\times H\times W\times 5}$; episode type $e$; patch size $p$;
atomic label set $\mathcal{A}=\{R,G,B,D,\tau\}$.
\ENSURE $(v^{\mathrm{mix}},\mathcal{C})=\textsc{RouteMix}(v^{(5)},e,p)$.
\STATE Initialize mixed clip $v^{\mathrm{mix}}\in\mathbb{R}^{T\times H\times W\times 3}$ and routing map $\mathcal{C}$.
\FOR{each spatial patch $(i,j)$}
    \STATE Sample a 3-element set $\mathcal{S}_{ij}\subset\mathcal{A}$ conditioned on $e$.
    \STATE Canonicalize to ordered slots:
    $(c_{ij,1},c_{ij,2},c_{ij,3})\leftarrow \mathrm{sort}(\mathcal{S}_{ij};\, D \succ \tau \succ R \succ G \succ B)$.
    \STATE \textbf{Temporal consistency:} reuse $(c_{ij,1},c_{ij,2},c_{ij,3})$ for all frames $t$ at location $(i,j)$.
    \STATE Route channels for all $t$:
    $v^{\mathrm{mix}}_t[i:i{+}p, j:j{+}p, s]\leftarrow v^{(5)}_t[i:i{+}p, j:j{+}p, c_{ij,s}]$ for $s\in\{1,2,3\}$.
    \STATE Store $\mathcal{C}_{ij}\leftarrow (c_{ij,1},c_{ij,2},c_{ij,3})$.
\ENDFOR
\RETURN $v^{\mathrm{mix}},\mathcal{C}$.
\end{algorithmic}
\end{algorithm}

\begin{algorithm}[t]
\caption{Episode-driven multimodal masked pretraining for ReMiX-MAE}
\label{alg:remix_pretrain}
\begin{algorithmic}[1]
\REQUIRE Synchronized clips $\{r_t\}_{t=1}^{T}$, $\{d_t\}_{t=1}^{T}$, $\{\tau_t\}_{t=1}^{T}$;
episode distribution $\pi(e)$; patch size $p$; MAE mask ratio $r_{\mathrm{mask}}$.
\FOR{each iteration}
    \STATE Form the 5-channel clip: $v^{(5)}_t \leftarrow \mathrm{concat}(r_t,d_t,\tau_t)$ for $t=1,\dots,T$.
    \STATE Sample episode: $e\sim\pi(e)$.
    \STATE Routing: $(v^{\mathrm{mix}},\mathcal{C}) \leftarrow \textsc{RouteMix}(v^{(5)},e,p)$ (Alg.~\ref{alg:routing_identity}).
    \STATE MAE mask (episode-independent): sample $\mathcal{M}\subset\mathcal{P}$ with ratio $r_{\mathrm{mask}}$; let $\mathcal{V}=\mathcal{P}\setminus\mathcal{M}$.
    \STATE Declared-missing metadata: derive an episode-dependent set $\mathcal{D}=\mathcal{D}(\mathcal{C},e)$ (does not alter $\mathcal{V}$).
    \STATE Form embeddings: $\tilde{E}\leftarrow \mathrm{PatchEmbed}(v^{\mathrm{mix}})+\mathrm{PosEmbed}+\mathrm{IdEmb}(\mathcal{C})$.
    \STATE Encode visible tokens: $z \leftarrow f_\theta(\tilde{E}_{\mathcal{V}})$.
    \STATE Decode and compute core recon loss on $\mathcal{M}$: $\mathcal{L}_{\text{recon}}(\mathcal{M})$.
    \STATE Compute auxiliary channel-drop supervision on $\mathcal{D}$: $\mathcal{L}_{\text{cdrop}}(\mathcal{D})$.
    \STATE Compute total loss:
    $\mathcal{L}_{\text{total}} \leftarrow \mathcal{L}_{\text{recon}}(\mathcal{M}) + \lambda_{\text{cd}}\mathcal{L}_{\text{cdrop}}(\mathcal{D}) + \lambda_{\text{temp}}\mathcal{L}_{\text{temp}} + \lambda_{\text{frame}}\mathcal{L}_{\text{frame}}$.
    \STATE Update parameters using $\mathcal{L}_{\text{total}}$.
\ENDFOR
\end{algorithmic}
\end{algorithm}
To preserve a 3-channel interface while leveraging five atomic modalities, we perform patch-level 5-to-3 routing.
Let $\mathcal{A}=\{R,G,B,D,\tau\}$ denote the atomic channel labels. We fix the channel order of $v^{(5)}$ as $[D,\tau,R,G,B]$,
so each label $c\in\mathcal{A}$ uniquely corresponds to its channel index in the 5-channel tensor.
For each spatial patch $(i,j)$, we sample a 3-element set
$\mathcal{S}_{ij}\subset\mathcal{A}$ and convert it into an ordered tuple using a fixed deterministic tie-breaker,
$(c_{ij,1},c_{ij,2},c_{ij,3})=\mathrm{sort}(\mathcal{S}_{ij};\, D \succ \tau \succ R \succ G \succ B)$.
We then form the mixed clip by copying the selected channels into slots $\ell\in\{1,2,3\}$:
\[
v^{\mathrm{mix}}_t[i:i{+}p,\,j:j{+}p,\,\ell] \;=\; v^{(5)}_t[i:i{+}p,\,j:j{+}p,\,c_{ij,\ell}].
\]
The routing tuple is kept fixed across time for each spatial location to ensure temporal consistency. The collection
$\mathcal{C}=\{c_{ij,\ell}\}$ defines the routing map (we refer to this procedure as \textsc{RouteMix}).

Routing yields mixed-channel tokens whose slot provenance varies across tokens. To make this provenance explicit to the encoder,
we inject a token-wise modality-identity bias. Let $(c_{k,1},c_{k,2},c_{k,3})\in\mathcal{A}^3$ denote the ordered routing labels
associated with token $k\in\mathcal{P}$. We learn an embedding table $u:\mathcal{A}\rightarrow\mathbb{R}^{d}$ and aggregate the
three slot identities via a shared DeepSets-style encoder:
\[
s_k=\sum_{s=1}^{3}\phi\!\big(u(c_{k,s})\big)\in\mathbb{R}^{d},
\qquad
b_k=\mathrm{LN}\!\big(\rho(s_k)\big)\in\mathbb{R}^{d},
\]
where $\phi(\cdot)$ is a shared MLP, $\rho(\cdot)$ is a linear projection to the token
embedding dimension $d$, and $\mathrm{LN}$ denotes LayerNorm. Stacking over tokens defines the identity embedding matrix
\[
\mathrm{IdEmb}(\mathcal{C}) \;\triangleq\; [\,b_1;\,b_2;\,\dots;\,b_N\,]\in\mathbb{R}^{N\times d}.
\]
Finally, we form encoder input embeddings by combining patch embeddings, positional encoding, and the identity bias:
\[
\tilde{E}=\mathrm{PatchEmbed}(v^{\mathrm{mix}})+\mathrm{PosEmbed}+\mathrm{IdEmb}(\mathcal{C}).
\]

\subsection{Multimodal Masked Pretraining with Episode-Based Missing-Modality Simulation}
\label{subsec:pretrain_objective}

Let $\mathcal{P}=\{1,\dots,N\}$ denote the token index set. We sample an MAE mask $\mathcal{M}\subset\mathcal{P}$ and define the visible set $\mathcal{V}=\mathcal{P}\setminus\mathcal{M}$. The encoder processes only the visible tokens $\tilde{E}_{\mathcal{V}}$. Decoders produce predictions on the full token grid, and reconstruction losses are computed by indexing masked locations (e.g., $(\cdot)_{\mathcal{M}}$), together with auxiliary terms described below.

\subsubsection{Core token-level reconstruction}
\label{subsec:core_recon}
We use modality-specific reconstruction forms.
For RGB, we regress patchified pixel values with an $\ell_2$ loss:
\begin{align*}
\mathcal{L}_{\mathrm{rgb}}(\mathcal{M})
&= \mathrm{MSE}\!\left(\hat{v}^{(\mathrm{rgb})}_{\mathcal{M}},\, v^{(\mathrm{rgb})}_{\mathcal{M}}\right).
\end{align*}

For thermal and depth, the decoder heads predict probabilistic outputs parameterized as Gaussians with mean $\hat{\mu}$ and log-variance $\hat{s}=\log\hat{\sigma}^2$. We optimize the Gaussian negative log-likelihood (NLL) on masked tokens:
\begin{equation}
\mathcal{L}_{\mathrm{nll}}(\hat{\mu},\hat{s};v)
=\tfrac{1}{2}\sum_{i\in\mathcal{M}}\left(\hat{s}_i+\exp(-\hat{s}_i)\,(v_i-\hat{\mu}_i)^2\right).
\end{equation}

For depth, we additionally employ a scale-invariant loss $\ell_{\mathrm{si}}$ on the predicted mean $\hat{\mu}^{(\mathrm{d})}$ to emphasize relative geometry:
\begin{equation}
r_i=\hat{\mu}^{(\mathrm{d})}_i-v^{(\mathrm{d})}_i,\qquad
\ell_{\mathrm{si}}=\sqrt{\frac{1}{n}\sum_{i\in\mathcal{M}} r_i^2-\frac{1}{n^2}\Big(\sum_{i\in\mathcal{M}} r_i\Big)^2},
\end{equation}
where $r_i$ denotes the per-token residual and $n=|\mathcal{M}|$.

The total losses for thermal ($\tau$) and depth (d) combine a base data term with the uncertainty-aware NLL:
\begin{equation}
\mathcal{L}_{\tau}(\mathcal{M})
=\ell_{\text{char}}\!\left(\hat{\mu}^{(\tau)}_{\mathcal{M}},\,v^{(\tau)}_{\mathcal{M}}\right)
+\lambda_{\text{nll}}\,\mathcal{L}_{\mathrm{nll}}\!\left(\hat{\mu}^{(\tau)}_{\mathcal{M}},\hat{s}^{(\tau)}_{\mathcal{M}};\,v^{(\tau)}_{\mathcal{M}}\right),
\end{equation}
\begin{equation}
\mathcal{L}_{\mathrm{d}}(\mathcal{M})
=\ell_{\mathrm{si}}\!\left(\hat{\mu}^{(\mathrm{d})}_{\mathcal{M}},\,v^{(\mathrm{d})}_{\mathcal{M}}\right)
+\lambda_{\text{nll}}\,\mathcal{L}_{\mathrm{nll}}\!\left(\hat{\mu}^{(\mathrm{d})}_{\mathcal{M}},\hat{s}^{(\mathrm{d})}_{\mathcal{M}};\,v^{(\mathrm{d})}_{\mathcal{M}}\right),
\end{equation}
where $\ell_{\text{char}}$ denotes the Charbonnier loss.

We combine modality-specific losses as:
\begin{equation}
\mathcal{L}_{\text{recon}}(\mathcal{M})
=
\alpha_{\mathrm{rgb}}\mathcal{L}_{\mathrm{rgb}}(\mathcal{M})
+
\alpha_{\tau}\mathcal{L}_{\tau}(\mathcal{M})
+
\alpha_{\mathrm{d}}\mathcal{L}_{\mathrm{d}}(\mathcal{M}).
\label{eq:loss_recon}
\end{equation}

\subsubsection{Overall objective}
\label{subsec:overall_objective}

We optimize a weighted sum of the core reconstruction loss and auxiliary terms:
\begin{equation}
\mathcal{L}_{\text{total}}
=
\mathcal{L}_{\text{recon}}(\mathcal{M})
+
\lambda_{\text{cd}}\,\mathcal{L}_{\text{cdrop}}(\mathcal{D})
+
\lambda_{\text{temp}}\,\mathcal{L}_{\text{temp}}
+
\lambda_{\text{frame}}\,\mathcal{L}_{\text{frame}}.
\end{equation}
Here, $\mathcal{L}_{\text{cdrop}}(\mathcal{D})$ reuses the modality-specific reconstruction objectives in Eq.~\eqref{eq:loss_recon} but applies them to the episode-declared missing modalities $\mathcal{D}$, enforcing signal completion when the corresponding input channels are explicitly dropped.
$\mathcal{L}_{\text{temp}}$ and $\mathcal{L}_{\text{frame}}$ are optional temporal and frame-level regularizers on reconstructed videos; we provide their exact forms in the supplementary material.

\subsection{Pretraining Data and Transfer to RGB-Only Clinical Videos}
\label{subsec:pretrain_transfer}
We pretrain ReMiX-MAE on BP4D+~\cite{bp4d+}, a large-scale facial behavior dataset that provides synchronized RGB, thermal, and depth recordings but is not pain-related. We choose BP4D+ because it offers well-aligned multimodal facial videos at scale, making it suitable for learning cross-modal facial representations without introducing pain-specific supervision. Pretraining is fully self-supervised: we use only raw videos and do not use any BP4D+ labels.

Downstream pain datasets provide RGB-only videos. We deploy the pretrained model in two RGB-only modes: \textbf{ReMiX-RGB} uses the frozen encoder as a feature extractor with routing constrained to RGB channels; \textbf{ReMiX-Pseudo} uses the decoders to reconstruct depth/thermal from RGB and then extracts features from a pseudo-multimodal, channel-mixed input. Both modes respect the same deployment constraint (RGB-only input) and enable a controlled study of whether pseudo-multimodal reconstruction provides additional benefit.

\section{Experiments}
\label{sec:experiments}

We evaluate ReMiX along four axes: clinical performance on SMP, cross-database transfer to UNBC, label efficiency on BioVid against TNT under limited supervision, and mechanism ablations. We additionally report an exploratory longitudinal analysis enabled by SMP's paired pre/post design.


\subsection{Experimental Protocols}
\label{subsec:protocols}

\begin{table}[t]
\centering
\small
\setlength{\tabcolsep}{6pt}
\caption{Dataset overview. This count reflects frames retained after face detection and quality filtering.}
\label{tab:dataset_overview}
\begin{tabular}{lcccc}
\toprule
Dataset & \#Subj. & \#Videos & Total \#Frames & Label \\
\midrule
SMP (ours) & 36 & 72 (1 min) & $\sim 110{,}000$ & NRS\\
UNBC & 25 & 200 (5s--30s) & $\sim 50{,}000$ & VAS\\
BioVid & 86 & 8600 (5s) & $\sim 1{,}000{,}000$ & Stimulus\\
\bottomrule
\end{tabular}
\vspace{-2mm}
\end{table}

\subsubsection{Datasets and task formulation}
We evaluate on three datasets: SMP (ours), UNBC~\cite{ShoulderPain}, and BioVid~\cite{walter2013biovid}. Table~\ref{tab:dataset_overview} summarizes their statistics and label definitions.
On SMP, each recording has a video-level self-report pain score; we cast pain recognition as ordinal classification by discretizing scores into $K\in\{3,4,5\}$ bins (primary: $K=5$; others for reference). Bin definitions and class distributions are fixed across all experiments and provided in the supplementary material.

For feature extraction, each video is segmented into 5-second clips. Within each clip, we extract features from multiple non-overlapping temporal windows and aggregate window-level features to predict clip-level labels, yielding weak supervision at the clip level.

\subsubsection{Frozen representations}
\begin{figure}[t]
  \centering
  \includegraphics[width=\linewidth]{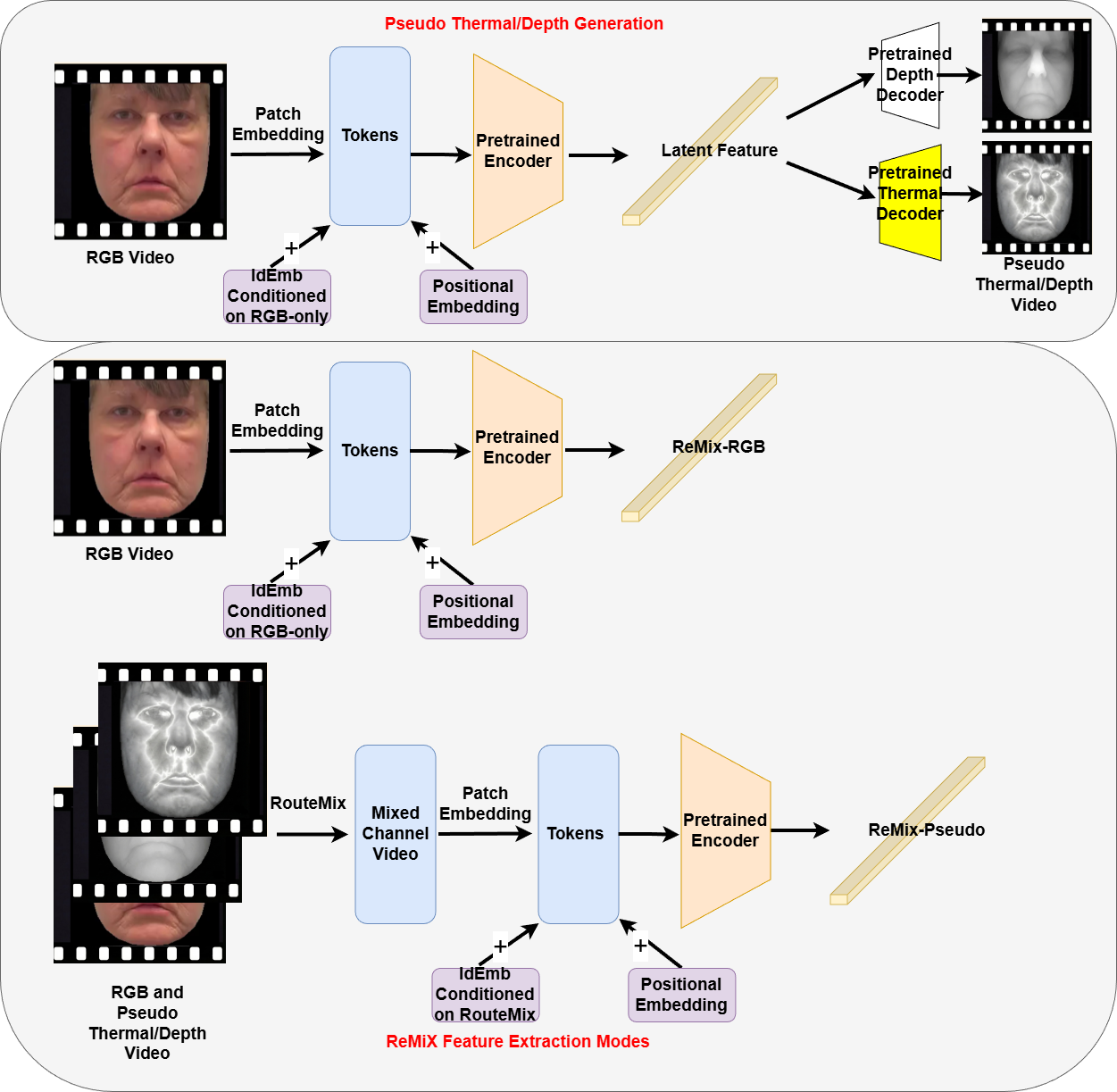}
  \caption{ReMiX feature extraction at RGB-only deployment.}
  \label{fig:remix_rgb_pseudo}
\end{figure}

We compare three frozen feature families under a unified downstream protocol:
\emph{(i) Baseline(AU)}, engineered facial features from OpenFace (17 AU intensities plus PSPI);
\emph{(ii) Baseline(MAE)}, RGB-only self-supervised representations extracted using MARLIN~\cite{cai2023marlin} pretrained on YouTube Faces~\cite{ytbface} and further trained on unlabeled BP4D+ videos for fair comparison; and
\emph{(iii) ReMiX}, features extracted using ReMiX-MAE under the two RGB-only deployment modes defined in Sec.~\ref{subsec:pretrain_transfer} (Fig.~\ref{fig:remix_rgb_pseudo}).
All encoders are kept frozen throughout downstream training.

\subsubsection{Downstream learning and augmentation}
We aggregate window-level features within each clip using a Multiple Instance Learning (MIL) module~\cite{ilse2018attention,dietterich1997solving} and train a lightweight classifier on top of frozen representations following common representation-evaluation practice~\cite{chen2020simple}. SMP experiments use leave-one-subject-out (LOSO) cross-validation. We report both clip-level metrics and video-level metrics obtained by majority voting over clip predictions; cross-database evaluations reuse the same downstream components unless otherwise noted.

Face-swap augmentation~\cite{simswapplusplus} is applied only to training videos within each LOSO fold. Donor identities are sampled from an external, non-overlapping face set unrelated to SMP/UNBC/BioVid; the held-out test subject is never augmented, and no evaluation identities are used as donors, preventing identity leakage across folds. Validation and test videos are always kept in their original, non-augmented form. We sweep the augmentation ratio and use 0.75 in all paired and cross-database analyses; full sweep results are provided in the supplementary material. These design choices (LOSO and identity-swap) help reduce subject-specific bias in small-cohort clinical learning.

\subsection{Clinical Performance on SMP under RGB-only Deployment}
\label{subsec:e1-syracuse}

\begin{table}[t]
\centering
\footnotesize
\setlength{\tabcolsep}{2.2pt}
\caption{Frozen-feature comparisons on SMP under LOSO. Subject-level and video-level metrics are reported; best results are in bold.}
\label{tab:syracuse_main}
\begin{threeparttable}
\begin{tabular}{llcccc}
\toprule
& & \multicolumn{2}{c}{Subject-level} & \multicolumn{2}{c}{Video-level} \\
\cmidrule(lr){3-4} \cmidrule(lr){5-6}
$K$ & Feature Setting & Acc. ($\pm$CI) & Macro-F1 ($\pm$CI) & Top-1 & Top-2 \\
\midrule
3 & Baseline (AU)          & 0.351 $\pm$ 0.032 & 0.226 $\pm$ 0.026 & \textbf{0.469} & 0.719 \\
3 & Baseline (MAE)         & 0.318 $\pm$ 0.050 & 0.196 $\pm$ 0.028 & 0.281 & 0.672 \\
3 & \textbf{Ours (RGB)}    & \textbf{0.398} $\pm$ 0.057 & \textbf{0.248} $\pm$ 0.040 & 0.453 & \textbf{0.812} \\
3 & \textbf{Ours (Pseudo)} & 0.394 $\pm$ 0.057 & 0.236 $\pm$ 0.031 & 0.391 & 0.719 \\
\midrule
4 & Baseline (AU)          & 0.265 $\pm$ 0.035 & 0.138 $\pm$ 0.021 & \textbf{0.375} & 0.656 \\
4 & Baseline (MAE)         & 0.290 $\pm$ 0.055 & 0.137 $\pm$ 0.019 & 0.219 & 0.594 \\
4 & \textbf{Ours (RGB)}    & 0.315 $\pm$ 0.054 & 0.153 $\pm$ 0.026 & 0.312 & \textbf{0.672} \\
4 & \textbf{Ours (Pseudo)} & \textbf{0.319} $\pm$ 0.059 & \textbf{0.156} $\pm$ 0.027 & 0.359 & 0.641 \\
\midrule
5 & Baseline (AU)          & 0.184 $\pm$ 0.046 & 0.082 $\pm$ 0.019 & 0.250 & 0.312 \\
5 & Baseline (MAE)         & 0.190 $\pm$ 0.040 & 0.082 $\pm$ 0.015 & 0.203 & 0.344 \\
5 & \textbf{Ours (RGB)}    & 0.237 $\pm$ 0.047 & 0.095 $\pm$ 0.023 & \textbf{0.297} & 0.469 \\
5 & \textbf{Ours (Pseudo)} & \textbf{0.255} $\pm$ 0.062 & \textbf{0.110} $\pm$ 0.024 & \textbf{0.297} & \textbf{0.500} \\
\bottomrule
\end{tabular}

\begin{tablenotes}[flushleft]
\footnotesize
\item \textbf{Note:} Baseline (MAE) uses MARLIN-RGB features; Baseline (AU) uses AU+PSPI. Ours (RGB/Pseudo) correspond to ReMiX-RGB / ReMiX-Pseudo.
\end{tablenotes}
\end{threeparttable}
\end{table}

Table~\ref{tab:syracuse_main} reports clinical pain recognition on SMP under LOSO across ordinal discretizations $K\in\{3,4,5\}$. 
Under \emph{RGB-only} deployment, ReMiX-RGB consistently improves subject-level Accuracy and Macro-F1 over the RGB-only self-supervised baseline (MARLIN-RGB) for all $K$, suggesting that multimodal masked pretraining enhanced representation quality under \emph{RGB-only} deployment for clinically grounded pain assessment.
ReMiX-Pseudo further incorporates decoder-generated pseudo thermal/depth cues with patch-wise channel routing under the same downstream protocol, providing additional gains over ReMiX-RGB, most notably in the finer-grained settings ($K=4,5$) where adjacent pain levels are harder to separate.

\begin{figure}[t]
\centering
\includegraphics[width=\columnwidth]{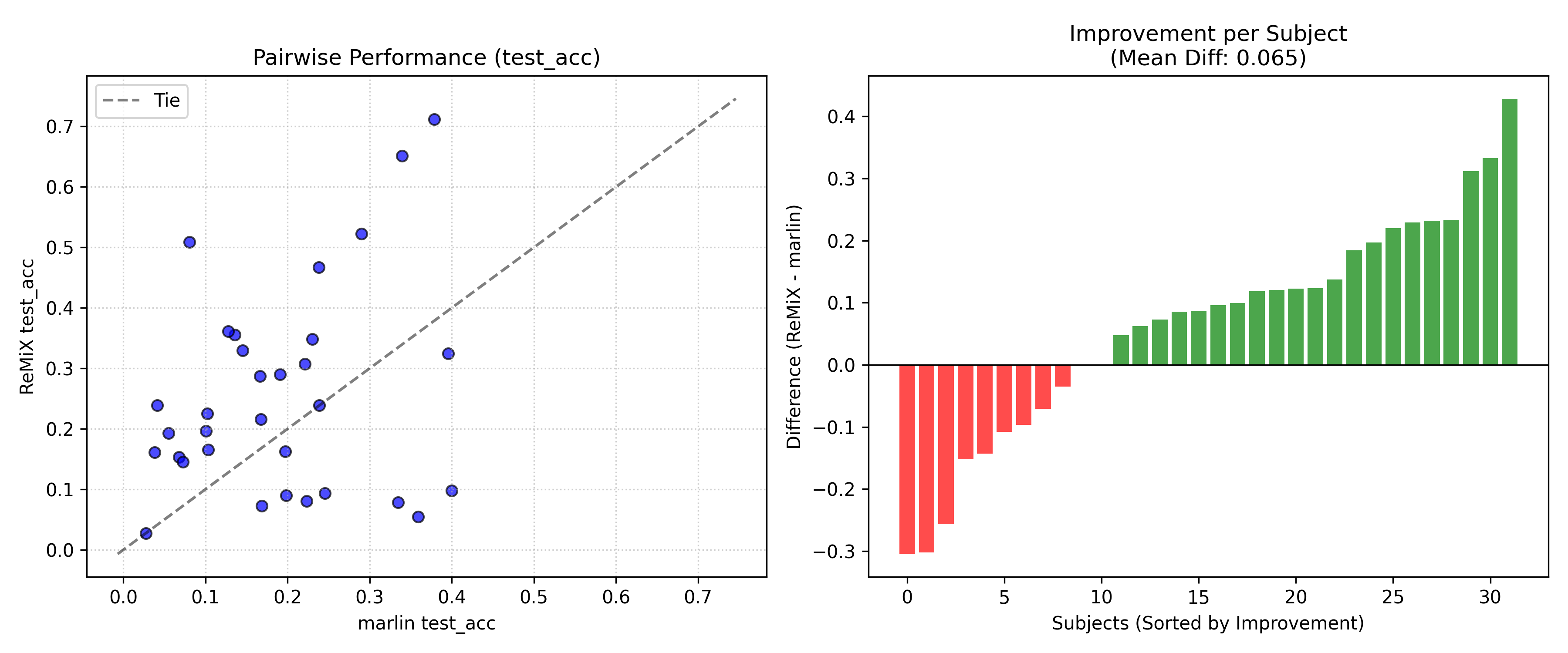}
\vspace{-2mm}
\caption{Representative subject-wise paired comparison on SMP under LOSO ($K=5$, augmentation ratio $=0.75$). Each point is a held-out subject; the right panel shows per-subject accuracy differences (Ours minus MARLIN).}
\label{fig:subjectwise_pairwise_rep}
\vspace{-2mm}
\end{figure}

To verify that the observed gains are not driven by a small subset of subjects, we report a representative subject-wise paired comparison for $K=5$ in Fig.~\ref{fig:subjectwise_pairwise_rep}; additional comparisons (including $K=3$ and the AU/PSPI baseline) are provided in the supplementary material.

As a diagnostic, we evaluate TNT~\cite{gkikas2023full} under the SMP LOSO protocol. Although it converges on BioVid, transfer to SMP with a conservative adaptation scheme (freezing pretrained modules and tuning only the classifier head) is near chance, underscoring the difficulty of direct temporal fitting under weak clip/video-level supervision.


\subsection{Cross-database Evaluation on UNBC}
\label{subsec:e3-crossdb}

\begin{table}[t]
\centering
\small
\setlength{\tabcolsep}{6pt}
\caption{Cross-database evaluation on UNBC (5-class; self-report pain score as ground truth). Best results per metric are in bold.}
\label{tab:crossdb_shoulder_pain}
\begin{tabular}{lcc}
\toprule
Method & Acc. & Macro-F1 \\
\midrule
Baseline (AU)          & 0.217 & 0.132 \\
Baseline (MAE)         & 0.172 & 0.103 \\
\textbf{Ours (RGB)}    & \textbf{0.280} & 0.132 \\
\textbf{Ours (Pseudo)} & 0.268 & \textbf{0.146} \\
\bottomrule
\end{tabular}
\vspace{-1mm}
\end{table}



For source-only transfer, we train on SMP and test on UNBC using five-class self-reported pain scores with the same cutoffs as SMP. No face-swap augmentation is used, and a small subject-disjoint UNBC subset is reserved for early stopping. Table~\ref{tab:crossdb_shoulder_pain} shows that, although all methods degrade under cross-database shift, ReMiX generalizes more favorably than the baselines: Ours (RGB) attains the highest accuracy and Ours (Pseudo) the best Macro-F1.

\subsection{Label-efficient Learning under Limited Supervision}
\label{biovid}

\begin{table}[t]
\centering
\small
\setlength{\tabcolsep}{5pt}
\caption{Data-efficiency on BioVid (Mean $\pm$ CI). Best results in bold.}
\label{tab:biovid_data_efficiency}
\begin{threeparttable}
\begin{tabular}{llcc}
\toprule
Ratio & Method & Acc. & Macro-F1 \\
\midrule
\multirow{2}{*}{25\%} & Baseline (TNT)\tnote{1} & 0.251 ± 0.015 & 0.199 ± 0.017 \\
                      & Ours (Pseudo)           & \textbf{0.260 ± 0.016} & \textbf{0.201 ± 0.016} \\
\midrule
\multirow{2}{*}{50\%} & Baseline (TNT)\tnote{1} & 0.256 ± 0.014 & 0.200 ± 0.017 \\
                      & Ours (Pseudo)           & \textbf{0.262 ± 0.015} & \textbf{0.211 ± 0.015} \\
\bottomrule
\end{tabular}
\begin{tablenotes}[flushleft, para]
\scriptsize
\item[1] TNT results are from our re-implementation under the unified protocol in Sec.~\ref{subsec:protocols}.
\end{tablenotes}
\end{threeparttable}
\vspace{-2mm}
\end{table}

BioVid provides a tightly controlled stimulus-driven benchmark at scale. Although its labels differ in meaning from clinical self-report, it offers a practical testbed for evaluating whether self-supervised pretrained representations can make better use of scarce labeled data than end-to-end supervised temporal training. We compare Ours (Pseudo) to a strong supervised temporal transformer baseline, the Transformer in Transformer (TNT)~\cite{gkikas2023full}, under reduced-data regimes (25\% and 50\% of the training split). As shown in Table~\ref{tab:biovid_data_efficiency}, Ours (Pseudo) achieves consistent gains over TNT at both ratios, supporting self-supervised pretraining as a label-efficient and clinically practical alternative when reliable annotations are limited.

\subsection{Ablation Studies}
\label{subsec:ablations_mechanisms}

\begin{table}[t]
\centering
\small
\setlength{\tabcolsep}{3pt}
\caption{Mechanism ablations on SMP under RGB-only deployment ($K{=}5$), ablating IdEmb and Episode with RouteMix fixed.}

\label{tab:ablations_mechanisms}
\begin{tabularx}{\columnwidth}{Xcc|cc}
\toprule
Method & IdEmb & Episode & Acc. & Macro-F1 \\
\midrule
Baseline(MAE) & -- & -- & 0.190 $\pm$ 0.040 & 0.082 $\pm$ 0.015 \\
Ours (full) & \checkmark & \checkmark & $0.237 \pm 0.047$ & $0.095 \pm 0.023$ \\
\quad w/o IdEmb & \xmark & \checkmark & $0.211 \pm 0.049$ & $0.085 \pm 0.033$ \\
\quad w/o Episode & \checkmark & \xmark & $0.229 \pm 0.052$ & $0.090 \pm 0.023$ \\
\bottomrule
\end{tabularx}
\end{table}

Table~\ref{tab:ablations_mechanisms} provides empirical justification for two key design choices under RGB-only deployment with RouteMix fixed. Removing modality-identity injection (IdEmb) degrades performance, indicating that explicit token-wise modality identity helps align mixed-channel inputs into a more transferable latent space. Disabling Episode reduces performance, indicating that simulating structured modality-missing scenarios during pretraining helps the encoder learn cross-modal representations that remain robust when only RGB is available at deployment.

\subsection{Exploratory Longitudinal Pain Reduction Analysis}
\label{subsec:pain_reduction}


SMP includes within-subject paired pre/post recordings at clinically meaningful treatment timepoints, enabling a preliminary test of whether model predictions reflect therapeutic change. We define significant reduction as $\Delta y = y_{\text{pre}}-y_{\text{post}}\ge 3$ and derive predictions from the frozen 5-class estimator using $\Delta\hat{c}=\hat{c}_{\text{pre}}-\hat{c}_{\text{post}}\ge 2$, without training an additional model. Table~\ref{tab:pain_reduction} shows higher agreement for our features than for the RGB-only baseline. We view this as a feasibility study; explicit joint pre/post modeling is left for future work.

\begin{table}[t]
\centering
\small
\setlength{\tabcolsep}{7pt}
\caption{Exploratory pain-reduction tracking on paired pre/post SMP videos (5-class estimator).}
\label{tab:pain_reduction}
\begin{tabular}{lcc}
\toprule
Features (5-class estimator) & Acc. & F1 \\
\midrule
Baseline (MARLIN) & 0.500 & 0.400 \\
Ours (ReMiX) & \textbf{0.667} & \textbf{0.625} \\
\bottomrule
\end{tabular}
\vspace{-2mm}
\end{table}

\section{Conclusion and Future Work}

In this paper, we target clinically grounded pain assessment where supervision is weak and facial evidence is subtle and heterogeneous; we introduce the SMP dataset and ReMiX-MAE, which leverages cross-modal self-supervised pretraining to improve RGB-only deployment, yielding stronger performance than two baselines on SMP, better robustness under transfer to UNBC, and competitive label-efficient learning on BioVid under limited supervision. Several limitations should also be noted. SMP is a relatively small single-site clinical cohort, and broader demographic coverage as well as, where feasible, multi-site validation will be necessary to establish stronger claims about generalizability. While ReMiX-MAE shows consistent improvements over the evaluated baselines, the absolute performance remains modest, and we do not provide formal hypothesis testing across all settings. In addition, our empirical comparisons focus primarily on frozen-feature protocols under limited clinical supervision; stronger temporal models and broader supervised baselines remain important directions for future study. Finally, the paired pre/post analysis presented here is exploratory and does not yet model treatment trajectories jointly over time. Future work will therefore expand evaluation to more diverse clinical populations and settings, investigate stronger temporal aggregation and selective fine-tuning strategies, incorporate uncertainty-aware prediction, and explore more adaptive routing mechanisms for variable modality availability in practical deployment.


\appendix

\section{NRS Binning Schemes and Class Distributions}
\label{sec:supp_bins}

In the main paper, we model SMP pain prediction as an ordinal $K$-class classification problem, where the original 0--10 self-reported numeric rating scale (NRS) scores are grouped into $K\in\{3,4,5\}$ bins. To avoid interrupting the main experimental flow while ensuring reproducibility, we report the exact bin definitions and class distributions here.

Let $y\in\{0,\dots,10\}$ denote the self-reported NRS for a visit-level recording after the AU/PSPI-based quality control described in Section~\ref{sec:supp_pspi}. All downstream experiments are conducted on the remaining 64 videos. For each $K$, we sort videos by $y$ and partition them into approximately equally populated, contiguous NRS intervals. The resulting ranges and class counts are summarized in Table~\ref{tab:syracuse_class_dist_supp}.

\begin{table}[t]
\centering
\caption{Class distributions for the 3-, 4-, and 5-class NRS binning schemes on the cleaned SMP dataset (64 videos). Each row shows the NRS range, number of videos, and proportion of the dataset for that class.}
\label{tab:syracuse_class_dist_supp}
\small
\setlength{\tabcolsep}{5pt}
\begin{tabular}{lccc}
\toprule
\multicolumn{4}{c}{3-class} \\
\midrule
Class & NRS range & \#Videos & Percentage \\
\midrule
C0 & 0.0--2.0  & 12 & 18.8\% \\
C1 & 3.0--5.0  & 24 & 37.5\% \\
C2 & 6.0--10.0 & 28 & 43.8\% \\
\midrule
\multicolumn{4}{c}{4-class} \\
\midrule
Class & NRS range & \#Videos & Percentage \\
\midrule
C0 & 0.0--1.0  &  5 &  7.8\% \\
C1 & 2.0--3.0  & 15 & 23.4\% \\
C2 & 4.0--6.0  & 21 & 32.8\% \\
C3 & 7.0--10.0 & 23 & 35.9\% \\
\midrule
\multicolumn{4}{c}{5-class} \\
\midrule
Class & NRS range & \#Videos & Percentage \\
\midrule
C0 & 0.0--1.0  &  5 &  7.8\% \\
C1 & 2.0--3.0  & 15 & 23.4\% \\
C2 & 4.0--5.0  & 16 & 25.0\% \\
C3 & 6.0--7.0  & 13 & 20.3\% \\
C4 & 8.0--10.0 & 15 & 23.4\% \\
\bottomrule
\end{tabular}
\end{table}

The binning is intentionally heuristic: the goal is to obtain reasonably balanced classes while preserving ordinal structure. In the main paper, we report results for all three granularities and focus discussion on the 5-class setting as a compromise between label resolution and statistical reliability.

\section{PSPI and AU-Based Analysis on the SMP Dataset}
\label{sec:supp_pspi}

To better understand the relationship between facial action units (AUs), PSPI scores, and self-reported pain levels in SMP, we conducted a per-frame AU/PSPI analysis.

\subsection{Pipeline and Per-Video Statistics}
For each video, we extract frame-wise AUs with OpenFace and retain only frames with \texttt{success=1} and \texttt{confidence $\geq 0.9$}. PSPI is computed using the standard formulation:
\[
\mathrm{PSPI}=\mathrm{AU4}+\max(\mathrm{AU6},\mathrm{AU7})+\max(\mathrm{AU9},\mathrm{AU10})+\mathrm{AU43}.
\]
Since OpenFace does not provide AU43 directly, we follow common practice and approximate $\mathrm{AU43}$ with the blink-related signal AU45. We apply a short temporal smoothing window (0.1\,s) to the PSPI trajectory.

From each smoothed PSPI trajectory, we derive:
\begin{itemize}
  \item mean and variance of PSPI;
  \item the proportion of frames above a global ``high-pain'' threshold $p_{\text{high}}$;
  \item intensity statistics such as maximum PSPI and the 95th/99th percentiles.
\end{itemize}

We compare within-video variance of PSPI to between-video variance across the dataset. With 0.1\,s smoothing, the average within-video variance is approximately 0.35, whereas the between-video variance is about 2.40 (ratio $\approx 0.15$). Using a slightly larger smoothing window (0.2\,s) yields similar results, indicating limited sensitivity to the smoothing scale. Overall, most recordings exhibit very small high-pain frame ratios (median close to 0), with pain-related peaks appearing as sparse events rather than sustained segments.

\subsection{Alignment with Self-Reported NRS Labels}
We examine how these PSPI-derived statistics relate to sequence-level NRS scores. Using 72 videos with numeric pain annotations, we observe:
\begin{itemize}
  \item a slightly negative correlation between mean PSPI and NRS ($\mathrm{corr} \approx -0.07$);
  \item a near-zero correlation between $p_{\text{high}}$ and NRS ($\approx 0.03$);
  \item no clear monotonic trend when grouping videos into low (NRS $\leq 3$), medium (3--6), and high (NRS $>6$) buckets.
\end{itemize}

Qualitative inspection suggests two common mismatch patterns: (i) sustained high PSPI despite low reported pain or negative treatment outcome, often dominated by talking or interaction artifacts that strongly activate AU4/AU7; and (ii) low PSPI despite high NRS, where facial expressions remain subtle or ambiguous. Based on a fixed thresholding rule over PSPI statistics, we flag and exclude 8 recordings with extreme disagreement between PSPI and NRS from downstream experiments (yielding the 64 videos used in the main paper).

\subsection{Implications}
This analysis yields two practical conclusions. First, PSPI is not a reliable surrogate for self-reported NRS in this clinical setting and is therefore used only for quality control and exploratory analysis rather than supervision. Second, high-PSPI peaks are sparse in time, supporting the modeling choice in the main paper of segmenting each recording into short clips that inherit a single NRS label.

\section{Upstream Multimodal Pretraining with ReMiX-MAE}
\label{sec:supp_upstream}

This section provides additional details on the upstream multimodal pretraining used to obtain the ReMiX-MAE encoder employed in downstream experiments.

\subsection{Backbone and Multimodal Decoders}
ReMiX-MAE builds on a 3D Vision Transformer backbone with tubelet patch embedding. Given a clip of $T$ aligned frames from synchronized RGB, depth, and thermal cameras, we form per-frame atomic channels $\{R,G,B,\mathrm{Depth},\mathrm{Thermal}\}$ and apply a 3D convolutional patch embedding to map spatio-temporal tubelets into tokens. Tokens are processed by a ViT-style encoder with fixed sinusoidal position encodings.

On top of the shared encoder, we attach three lightweight decoders for RGB, thermal, and depth. Each decoder has its own embedding dimension and depth and receives encoder tokens through a modality-specific linear projection. The RGB decoder outputs 3 reconstruction channels. The thermal and depth decoders output both a mean and a log-variance per pixel, enabling heteroscedastic Gaussian negative log-likelihood (NLL) losses.

A Modality Identity Encoder injects modality-aware bias into each token based on which atomic channels occupy its three input slots. It assigns a learnable embedding to each atomic modality, aggregates embeddings of modalities present in a patch via a DeepSets-style MLP, and adds the resulting vector to the corresponding token before the transformer encoder.

\begin{table*}[t]
\centering
\caption{ReMiX-MAE architecture used for upstream multimodal pretraining on BP4D+. Key encoder/decoder hyperparameters corresponding to the model used in downstream SMP experiments.}
\label{tab:remix_arch}
\small
\setlength{\tabcolsep}{6pt}
\begin{tabular}{ll}
\toprule
Component & Configuration \\
\midrule
Input clip & $T=16$ frames, $224\times224$, temporal stride $=2$ \\
Patch / tubelet size & $16\times16$ spatial, 2 frames temporal \\
\midrule
Encoder backbone & ViT-style 3D encoder \\
\quad Embedding dim & 768 \\
\quad Depth / heads & 12 layers, 12 heads \\
\quad MLP ratio & 4.0, with $qkv\_bias$ and LayerNorm \\
\quad Positional encoding & Fixed sinusoidal 3D \\
\quad Flash attention & Enabled (\texttt{encoder\_use\_flash\_attn=true}) \\
\midrule
RGB decoder & 3-channel reconstruction; dim 384; 4 layers, 6 heads \\
Thermal decoder & mean + log-variance; dim 256; 4 layers, 4 heads \\
Depth decoder & mean + log-variance; dim 256; 4 layers, 4 heads \\
\midrule
Modality identity encoder & 5 atomic embeddings (R,G,B,Depth,Thermal); DeepSets-style MLP; additive token bias \\
\bottomrule
\end{tabular}
\end{table*}

\subsection{Patch-wise 5-to-3 Channel Routing}
Instead of feeding modalities as separate streams, ReMiX-MAE performs early fusion via patch-wise channel routing. For each spatial patch, we sample three distinct channels from the five atomic channels and assign them to three input slots. The same triplet is reused across all frames at that spatial location to preserve temporal consistency. A deterministic priority rule (Depth $\succ$ Thermal $\succ$ R $\succ$ G $\succ$ B) reorders channels so that slot 0/1/2 have stable semantics.

This 5-to-3 routing exposes the encoder to diverse local modality combinations (e.g., RGB+Depth, RGB+Thermal, Depth+Thermal) without increasing the number of input channels beyond three. In \texttt{rgb\_only} episodes, routing is constrained to select the RGB channels, matching RGB-only deployment; in other episodes, mixed triplets are sampled to encourage multimodal fusion and robustness to locally missing cues.

\subsection{Episode Types}
Pretraining uses four episode types that share the same backbone/decoders but differ in routing constraints and auxiliary supervision:
\begin{itemize}
  \item \textbf{\texttt{mae}}: standard masked autoencoding with stochastic 5-to-3 routing and patch masking.
  \item \textbf{\texttt{rgb\_only}}: routing constrained to RGB; auxiliary completion emphasizes thermal/depth targets.
  \item \textbf{\texttt{cross\_modal}}: routing constrained to specific RGB/auxiliary combinations (details below).
  \item \textbf{\texttt{channel\_drop}}: modalities can be declared missing to simulate heterogeneous sensor failures; auxiliary completion is applied on the declared-missing targets.
\end{itemize}

\subsection{Cross-modal Episode Combinations}
\label{sec:supp_crossmodal_combos}
In \texttt{cross\_modal} episodes, we explicitly control the 3-slot input composition by sampling from a fixed set of RGB/auxiliary channel triplets. Each episode uniformly selects one triplet:
\[
\{\texttt{RGD},\,\texttt{RGT},\,\texttt{RDT},\,\texttt{GDT},\,\texttt{BDT}\},
\]
where, for example, \texttt{RGD} denotes (Red, Green, Depth) and \texttt{BDT} denotes (Blue, Depth, Thermal).\footnote{Channel indices in our implementation are: 0=R, 1=G, 2=B, 3=Depth, 4=Thermal.}
We select these five specific triplets to ensure a balanced coverage of cross-modal
interactions while avoiding redundancy. Specifically, each chosen triplet contains
at least one auxiliary channel (Depth or Thermal), and both auxiliary modalities
appear with equal frequency ($80\%$) to prevent modality bias. This selection
explicitly excludes the pure-RGB triplet $\{RGB\}$, which is already addressed
by the \texttt{rgb\_only} episode type, thereby forcing the encoder to focus
on learning correlations between geometric, physiological, and chromatic cues.

\subsection{Distinguishing the MAE Mask $\mathcal{M}$ from the Declared-missing Set $\mathcal{D}$}
\label{sec:supp_M_vs_D}
We use two independent notions of ``missingness'' during pretraining.

\textbf{Spatial MAE mask $\mathcal{M}$.} $\mathcal{M}\subset\mathcal{P}$ is a token-index set sampled by the MAE procedure and controls encoder visibility: tokens in $\mathcal{V}=\mathcal{P}\setminus\mathcal{M}$ enter the encoder, while tokens in $\mathcal{M}$ are removed and reconstructed by the decoder.

\textbf{Declared-missing set $\mathcal{D}$.} $\mathcal{D}$ is specified by the sampled episode type and indicates which modalities are treated as logically unavailable for auxiliary completion supervision (e.g., simulating sensor failure). When a modality is declared missing, its input signal is replaced with a learnable ``missing'' embedding (or zeros) without changing the token sequence length. Therefore, $\mathcal{D}$ does not alter which tokens enter the encoder; it only modifies the content of affected channels and the auxiliary loss indexing.

Crucially, $\mathcal{D}$ is independent of $\mathcal{M}$ and need not satisfy $\mathcal{D}\subseteq\mathcal{M}$. This allows tokens that are spatially visible but contain a declared-missing modality, so the encoder observes an explicit ``missing'' placeholder and must rely on contextual information, improving robustness under modality collapse or missing-sensor conditions.

\subsection{Pretraining Objective (Summary)}
The reconstruction loss combines modality-specific terms for RGB, thermal, and depth. For RGB, we use MSE on masked patches. For thermal and depth, we use a heteroscedastic Gaussian NLL and additional robust terms. Episode-specific declarations and weights determine where and how strongly the model is penalized for failing to reconstruct missing modalities, jointly encouraging cross-modal completion and missing-sensor compensation.

\subsection{Pretraining Data and Optimization}
The ReMiX-MAE encoder is obtained from self-supervised pretraining on BP4D+, which provides synchronized RGB, depth, and thermal facial videos recorded under controlled tasks. We initialize the encoder from a publicly available VideoMAE-style checkpoint trained on large-scale RGB video and continue training on BP4D+ using the mixed-modality, multi-episode objective described above.

Pretraining uses AdamW with cosine learning rate decay and linear warmup, mixed-precision training, and a high mask ratio on spatio-temporal tubelets. Episode types are sampled with fixed probabilities:
\[
\begin{aligned}
    p(\texttt{mae}) &= 0.40, \quad & p(\texttt{rgb\_only}) &= 0.40, \\
    p(\texttt{cross\_modal}) &= 0.15, \quad & p(\texttt{channel\_drop}) &= 0.05.
\end{aligned}
\]
After convergence, we freeze the encoder and reuse the same checkpoint for downstream experiments; only the downstream aggregation/classification modules are trained on clinical data.

\section{Effect of Identity-Swap Augmentation Ratio}
\label{sec:supp_aug_sweep}

To assess sensitivity to identity-swap augmentation, we sweep the augmentation ratio $\rho$ from 0 to 1.0 under the same LOSO protocol as the main experiments.
For each training video, we pre-generate an augmentation pool of $N_{\text{avail}}=4$ identity-swapped variants by swapping in four different donor identities sampled from an external, non-overlapping face set.
An augmentation ratio $\rho\in\{0,0.25,0.5,0.75,1.0\}$ determines how many variants are used per original training video in each epoch:
we set $N_{\text{use}}=\rho\,N_{\text{avail}}$ (i.e., $0/1/2/3/4$), and at each epoch we uniformly sample $N_{\text{use}}$ variants from the pool for each training video.
Thus, $\rho=0.25/0.5/0.75/1.0$ corresponds to sampling 1/2/3/4 augmented videos per original video per epoch, respectively.

\begin{figure}[t]
  \centering
  \includegraphics[width=\linewidth]{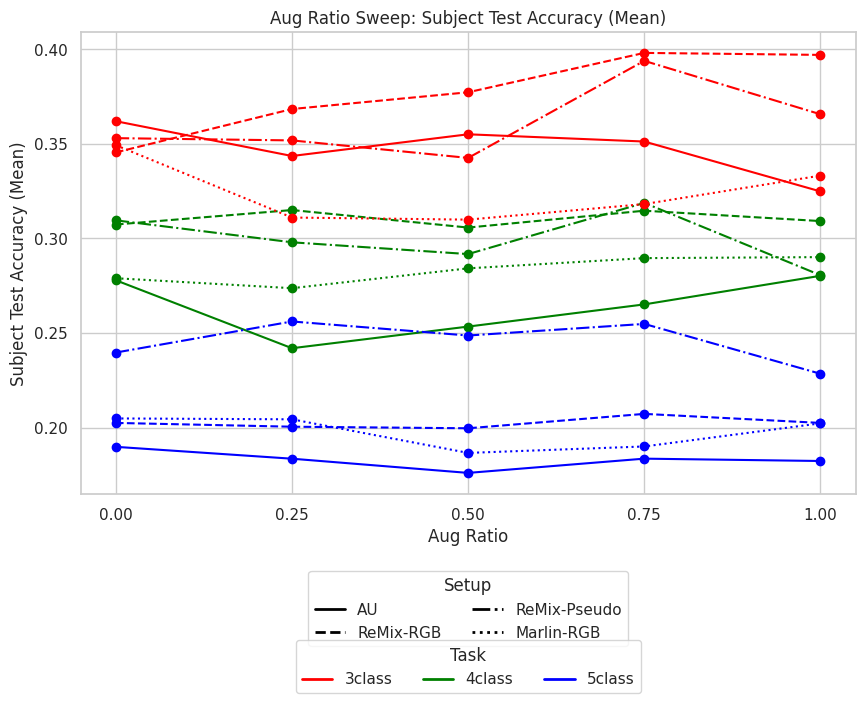}
  \caption{Subject-level test Accuracy under different augmentation ratios (mean over LOSO folds).}
  \label{fig:supp_aug_sweep_acc}
\end{figure}

\begin{figure}[t]
  \centering
  \includegraphics[width=\linewidth]{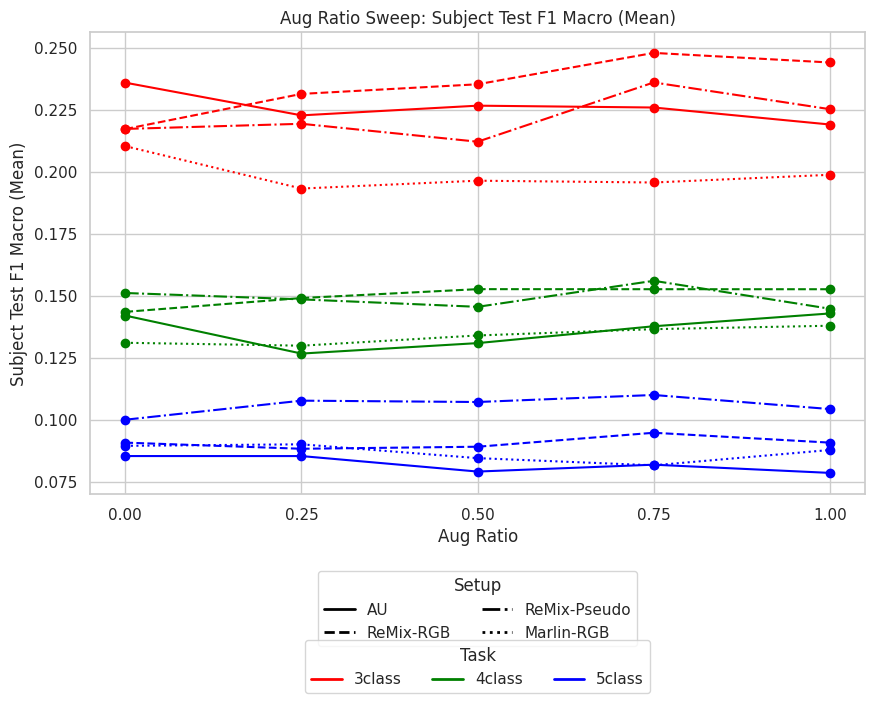}
  \caption{Subject-level test Macro-F1 under different augmentation ratios (mean over LOSO folds).}
  \label{fig:supp_aug_sweep_f1}
\end{figure}

Figure~\ref{fig:supp_aug_sweep_acc} and Figure~\ref{fig:supp_aug_sweep_f1} report the resulting subject-level test Accuracy and Macro-F1 (mean over LOSO folds) across $K\in\{3,4,5\}$ and all feature setups.

Overall, moderate augmentation ratios (approximately 0.5--0.75) yield the strongest performance, particularly for ReMiX-based representations. Extreme augmentation ($\rho=1.0$) does not provide further benefit and can slightly degrade performance, consistent with saturation or excessive synthetic variation. These trends motivate the shared augmentation ratio of 0.75 used in the main paper.

\section{Auxiliary Losses: Implementation Details}
\label{app:aux_losses}

\subsection{Declared-missing Channel-drop Loss}
Episodes define a declared-missing set $\mathcal{D}$ that specifies which modality targets are treated as missing for auxiliary completion supervision. In our implementation, $\mathcal{D}$ is represented by modality-specific token subsets $\mathcal{D}_\tau,\mathcal{D}_{\mathrm{d}}\subseteq\mathcal{P}$ for thermal and depth, respectively. We apply the same token-level loss forms as in the core reconstruction, but evaluated on $\mathcal{D}_\tau$ and $\mathcal{D}_{\mathrm{d}}$:
\[
\mathcal{L}_{\text{cdrop}}(\mathcal{D})
=
\ell_{\mathrm{charb}}\!\left(\hat{\mu}^{(\tau)}_{\mathcal{D}_\tau},\,x^{(\tau)}_{\mathcal{D}_\tau}\right)
+
\mathcal{L}_{\mathrm{nll}}\!\left(\hat{\mu}^{(\tau)}_{\mathcal{D}_\tau},\hat{s}^{(\tau)}_{\mathcal{D}_\tau};\,x^{(\tau)}_{\mathcal{D}_\tau}\right)
+
\ell_{\mathrm{si}}\!\left(\hat{\mu}^{(\mathrm{d})}_{\mathcal{D}_{\mathrm{d}}},\,x^{(\mathrm{d})}_{\mathcal{D}_{\mathrm{d}}}\right)
+
\mathcal{L}_{\mathrm{nll}}\!\left(\hat{\mu}^{(\mathrm{d})}_{\mathcal{D}_{\mathrm{d}}},\hat{s}^{(\mathrm{d})}_{\mathcal{D}_{\mathrm{d}}};\,x^{(\mathrm{d})}_{\mathcal{D}_{\mathrm{d}}}\right).
\]

\subsection{Frame-level Regularization $\mathcal{L}_{\text{frame}}$}
\label{sec:supp_Lframe}
We optionally apply additional regularizers on reconstructed frames and denote their aggregated contribution by $\mathcal{L}_{\text{frame}}$. In our implementation, $\mathcal{L}_{\text{frame}}$ includes (when enabled) an SSIM-based term and a temporal smoothness term on reconstructed thermal/depth videos:
(i) an SSIM-based loss $\mathcal{L}_{\text{ssim}}(I_{\text{pred}},I_{\text{gt}})$; and
(ii) a first-order temporal consistency penalty
\[
\mathcal{L}_{\text{temp}}=\frac{1}{T-1}\sum_{t=2}^{T}\left\|I_{\text{pred}}^{(t)}-I_{\text{pred}}^{(t-1)}\right\|_{1}.
\]
Each component is controlled by a scalar weight in the configuration. To avoid over-smoothing before basic structure is learned, we disable $\mathcal{L}_{\text{temp}}$ during the initial RGB warmup stage and activate it only after warmup.

\section{Subject-wise Paired Evaluation on SMP}
\label{sec:supp_subjectwise}

Aggregate improvements on small clinical datasets can be disproportionately influenced by a limited subset of subjects. To assess whether the gains reported in the main paper reflect broad subject-level trends rather than a few outliers, we perform subject-wise paired comparisons under the same LOSO protocol.

\subsection{Protocol}
For each LOSO fold, we evaluate ReMiX against each baseline on the held-out subject using identical downstream components (MIL aggregation and a shallow classification head) and the same training configuration. We report subject-level test accuracy for each held-out subject and compute paired differences between ReMiX and a given baseline on a per-subject basis. We present results for $K\in\{3,5\}$, corresponding to coarse and fine-grained clinical pain classification.

\subsection{Visualization and Interpretation}
\begin{figure*}[t]
\centering
\small
\begin{minipage}[t]{0.49\textwidth}
  \centering
  \includegraphics[width=\linewidth]{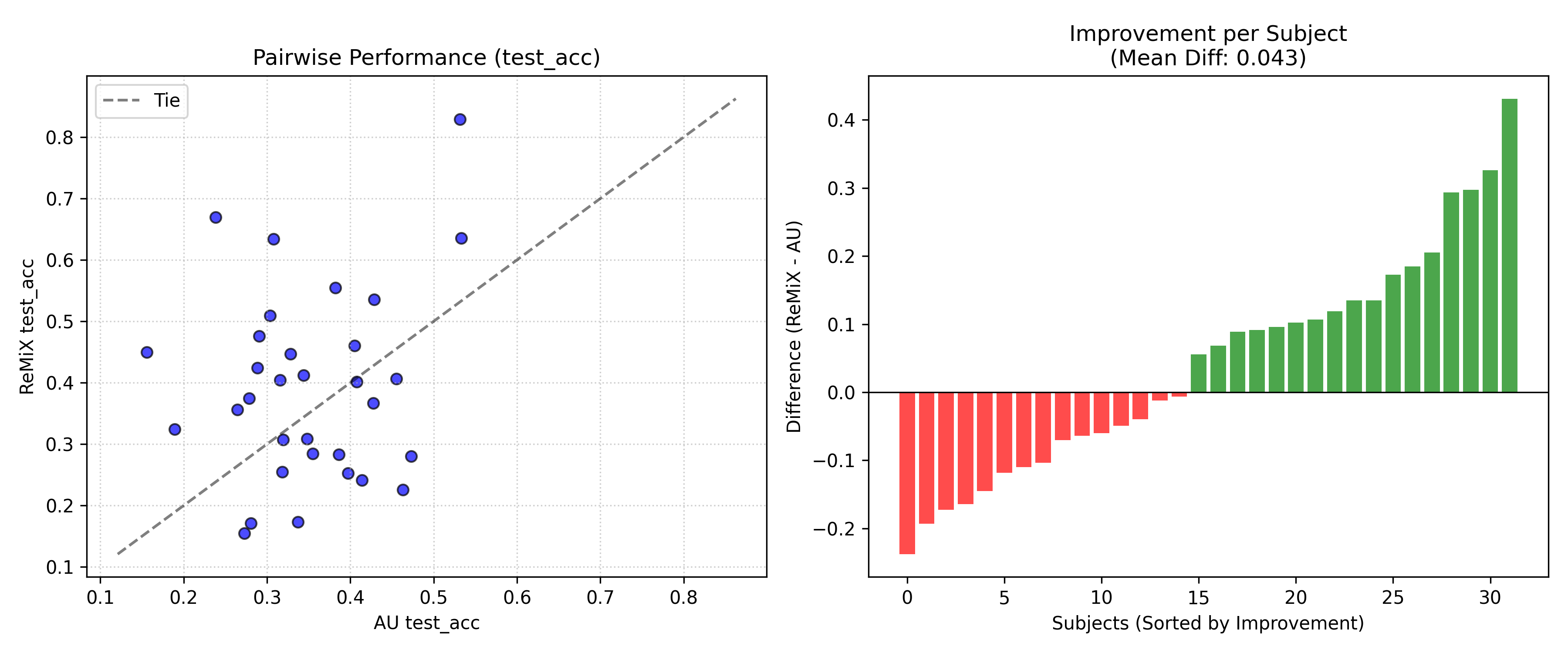}
  \vspace{-2mm}
  \par\small\textbf{(a) $K=3$: ReMiX vs.\ Baseline (AU)}
\end{minipage}
\hfill
\begin{minipage}[t]{0.49\textwidth}
  \centering
  \includegraphics[width=\linewidth]{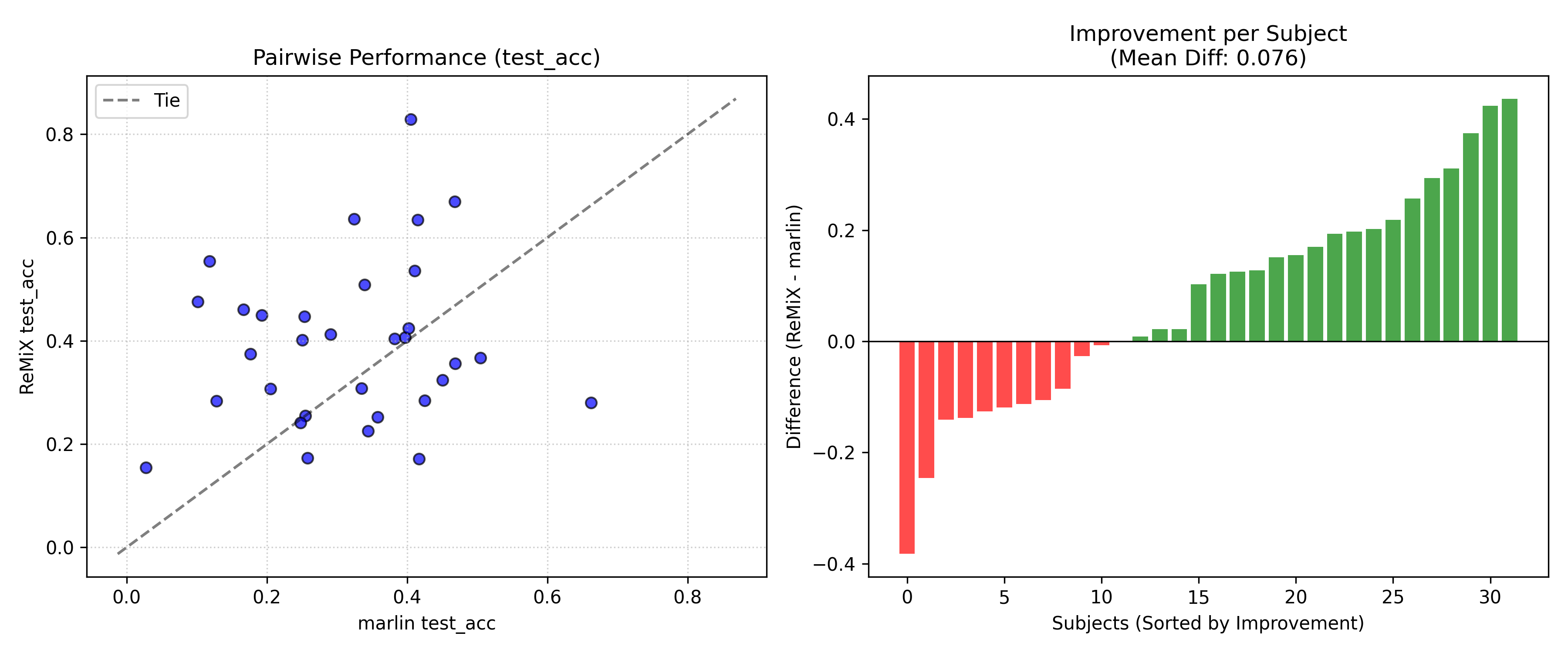}
  \vspace{-2mm}
  \par\small\textbf{(b) $K=3$: ReMiX vs.\ Baseline (MAE)}
\end{minipage}

\vspace{2mm}

\begin{minipage}[t]{0.49\textwidth}
  \centering
  \includegraphics[width=\linewidth]{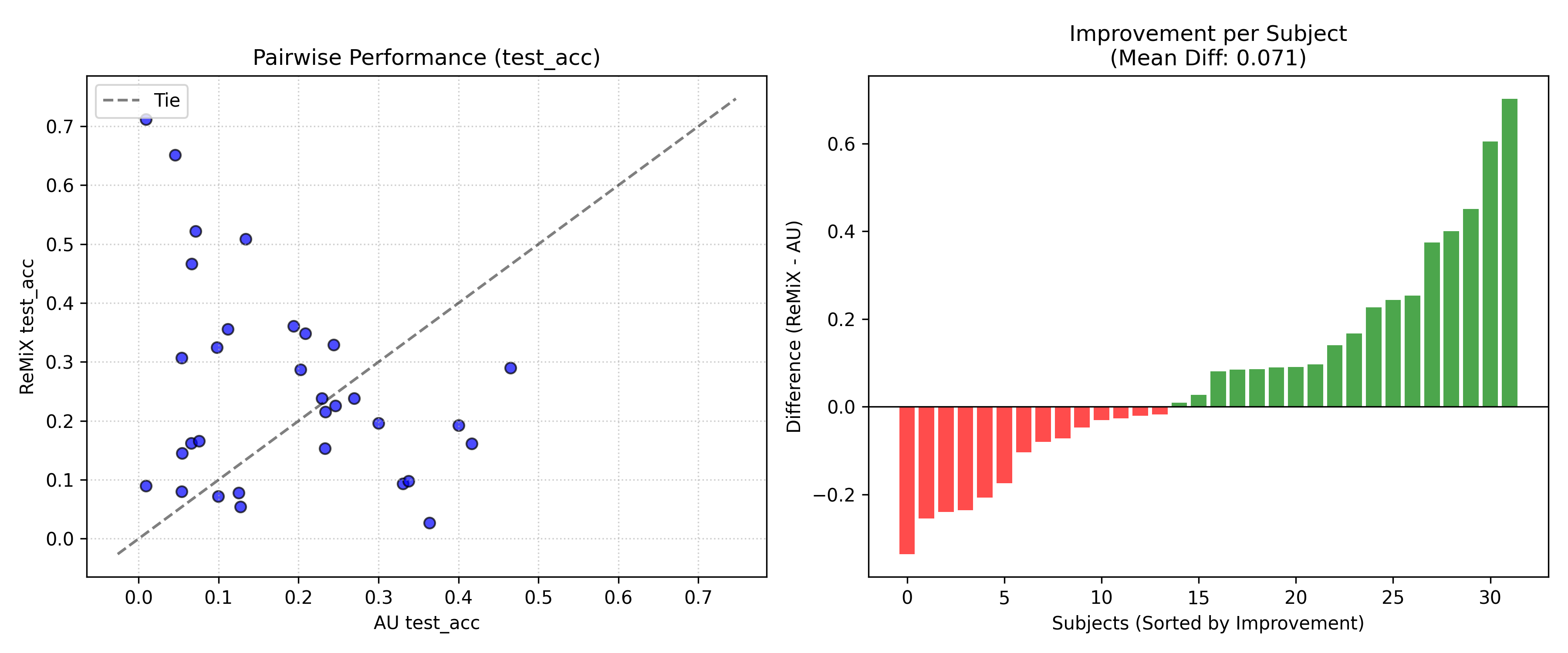}
  \vspace{-2mm}
  \par\small\textbf{(c) $K=5$: ReMiX vs.\ Baseline (AU)}
\end{minipage}
\hfill
\begin{minipage}[t]{0.49\textwidth}
  \centering
  \includegraphics[width=\linewidth]{5class_plot_marlin_vs_ReMiX_test_acc.png}
  \vspace{-2mm}
  \par\small\textbf{(d) $K=5$: ReMiX vs.\ Baseline (MAE)}
\end{minipage}

\caption{Subject-wise paired comparisons on SMP under LOSO (augmentation ratio $=0.75$). Left: subject-level test accuracies with the tie line. Right: sorted per-subject accuracy differences (ReMiX minus baseline). Mean Diff is the average of per-subject accuracy differences (equal weight per subject).}
\label{fig:supp_subjectwise_pairwise}
\vspace{-2mm}
\end{figure*}

Figure~\ref{fig:supp_subjectwise_pairwise} provides two complementary views: (1) a scatter plot of per-subject accuracies with the tie line; and (2) sorted per-subject accuracy differences (ReMiX minus baseline). Positive differences indicate improvements, while negative values reveal subject-dependent failure cases.

\subsection{Summary Statistic}
We summarize the overall tendency of subject-wise improvements using the mean paired difference (Mean Diff), computed as the average of per-subject accuracy differences (ReMiX minus baseline), giving equal weight to each held-out subject.

{\small
\bibliographystyle{ieee}
\bibliography{egbib}

@INPROCEEDINGS{ShoulderPain,
  author={Lucey, Patrick and Cohn, Jeffrey F. and Prkachin, Kenneth M. and Solomon, Patricia E. and Matthews, Iain},
  booktitle={2011 IEEE International Conference on Automatic Face \& Gesture Recognition (FG)}, 
  title={Painful data: The UNBC-McMaster shoulder pain expression archive database}, 
  year={2011},
  volume={},
  number={},
  pages={57-64},
  doi={10.1109/FG.2011.5771462}}

@inproceedings{walter2013biovid,
  title={The biovid heat pain database data for the advancement and systematic validation of an automated pain recognition system},
  author={Walter, Steffen and Gruss, Sascha and Ehleiter, Hagen and Tan, Junwen and Traue, Harald C and Werner, Philipp and Al-Hamadi, Ayoub and Crawcour, Stephen and Andrade, Adriano O and Da Silva, Gustavo Moreira},
  booktitle={2013 IEEE international conference on cybernetics (CYBCO)},
  pages={128--131},
  year={2013},
  organization={IEEE}
}

@inproceedings{werner2014automatic,
  title={Automatic pain recognition from video and biomedical signals},
  author={Werner, Philipp and Al-Hamadi, Ayoub and Niese, Robert and Walter, Steffen and Gruss, Sascha and Traue, Harald C},
  booktitle={2014 22nd international conference on pattern recognition},
  pages={4582--4587},
  year={2014},
  organization={IEEE}
}

@inproceedings{prajod2024faces,
  title={Faces of Experimental Pain: Transferability of Deep-Learned Heat Pain Features to Electrical Pain},
  author={Prajod, Pooja and Schiller, Dominik and Don, Daksitha Withanage and Andr{\'e}, Elisabeth},
  booktitle={2024 12th International Conference on Affective Computing and Intelligent Interaction Workshops and Demos (ACIIW)},
  pages={31--38},
  year={2024},
  organization={IEEE}
}

@inproceedings{prajod2022using,
  title={Using explainable AI to identify differences between clinical and experimental pain detection models based on facial expressions},
  author={Prajod, Pooja and Huber, Tobias and Andr{\'e}, Elisabeth},
  booktitle={International Conference on Multimedia Modeling},
  pages={311--322},
  year={2022},
  organization={Springer}
}

@article{emopain,
  title={The automatic detection of chronic pain-related expression: requirements, challenges and the multimodal EmoPain dataset},
  author={Aung, Min SH and Kaltwang, Sebastian and Romera-Paredes, Bernardino and Martinez, Brais and Singh, Aneesha and Cella, Matteo and Valstar, Michel and Meng, Hongying and Kemp, Andrew and Shafizadeh, Moshen and others},
  journal={IEEE transactions on affective computing},
  volume={7},
  number={4},
  pages={435--451},
  year={2015},
  publisher={IEEE}
}

@inproceedings{ai4pain,
  title={The ai4pain grand challenge 2025: Advancing pain assessment with multimodal physiological signals},
  author={Fernandez-Rojas, Raul and Joseph, Calvin and Hirachan, Niraj and Seymour, Ben and Goecke, Roland},
  booktitle={Companion Proceedings of the 27th International Conference on Multimodal Interaction},
  pages={147--152},
  year={2025}
}

@inproceedings{liu2018clinical,
  title={Clinical valid pain database with biomarker and visual information for pain level analysis},
  author={Liu, Peng and Yazgan, Idris and Olsen, Sarah and Moser, Alecia and Ciftci, Umur and Bajwa, Saeed and Tvetenstrand, Christian and Gerhardstein, Peter and Sadik, Omowunmi and Yin, Lijun},
  booktitle={2018 13th IEEE International Conference on Automatic Face \& Gesture Recognition (FG 2018)},
  pages={525--529},
  year={2018},
  organization={IEEE}
}

@inproceedings{zhang2024multimodal,
  title={Multimodal channel-mixing: Channel and spatial masked autoencoder on facial action unit detection},
  author={Zhang, Xiang and Yang, Huiyuan and Wang, Taoyue and Li, Xiaotian and Yin, Lijun},
  booktitle={Proceedings of the IEEE/CVF Winter Conference on Applications of Computer Vision},
  pages={6077--6086},
  year={2024}
}

@inproceedings{cai2023marlin,
  title={Marlin: Masked autoencoder for facial video representation learning},
  author={Cai, Zhixi and Ghosh, Shreya and Stefanov, Kalin and Dhall, Abhinav and Cai, Jianfei and Rezatofighi, Hamid and Haffari, Reza and Hayat, Munawar},
  booktitle={Proceedings of the IEEE/CVF conference on computer vision and pattern recognition},
  pages={1493--1504},
  year={2023}
}

@inproceedings{bp4d+,
  title={Multimodal spontaneous emotion corpus for human behavior analysis},
  author={Zhang, Zheng and Girard, Jeff M and Wu, Yue and Zhang, Xing and Liu, Peng and Ciftci, Umur and Canavan, Shaun and Reale, Michael and Horowitz, Andy and Yang, Huiyuan and others},
  booktitle={Proceedings of the IEEE conference on computer vision and pattern recognition},
  pages={3438--3446},
  year={2016}
}

@inproceedings{tong2022videomae,
  title={Video{MAE}: Masked Autoencoders are Data-Efficient Learners for Self-Supervised Video Pre-Training},
  author={Zhan Tong and Yibing Song and Jue Wang and Limin Wang},
  booktitle={Advances in Neural Information Processing Systems},
  year={2022}
}

@Article{simswapplusplus,
    author  = {Xuanhong Chen and
              Bingbing Ni and
              Yutian Liu and
              Naiyuan Liu and
              Zhilin Zeng and
              Hang Wang},
    title   = {SimSwap++: Towards Faster and High-Quality Identity Swapping},
    journal = {{IEEE} Trans. Pattern Anal. Mach. Intell.},
    volume  = {46},
    number  = {1},
    pages   = {576--592},
    year    = {2024}
}

@article{prkachin2008pspi,
  title={The structure, reliability and validity of pain expression: Evidence from patients with shoulder pain},
  author={Prkachin, Kenneth M and Solomon, Patricia E},
  journal={Pain},
  volume={139},
  number={2},
  pages={267--274},
  year={2008},
  publisher={Elsevier}
}

@article{werner_pain_survey,
  title={Automatic recognition methods supporting pain assessment: A survey},
  author={Werner, Philipp and Lopez-Martinez, Daniel and Walter, Steffen and Al-Hamadi, Ayoub and Gruss, Sascha and Picard, Rosalind W},
  journal={IEEE Transactions on Affective Computing},
  volume={13},
  number={1},
  pages={530--552},
  year={2019},
  publisher={IEEE}
}

@inproceedings{gkikas2023full,
  title={A full transformer-based framework for automatic pain estimation using videos},
  author={Gkikas, Stefanos and Tsiknakis, Manolis},
  booktitle={2023 45th Annual International Conference of the IEEE Engineering in Medicine \& Biology Society (EMBC)},
  pages={1--6},
  year={2023},
  organization={IEEE}
}

@inproceedings{zhou2016recurrent,
  title={Recurrent convolutional neural network regression for continuous pain intensity estimation in video},
  author={Zhou, Jing and Hong, Xiaopeng and Su, Fei and Zhao, Guoying},
  booktitle={Proceedings of the IEEE conference on computer vision and pattern recognition workshops},
  pages={84--92},
  year={2016}
}

@article{deeppain,
  title={Deep pain: Exploiting long short-term memory networks for facial expression classification},
  author={Rodriguez, Pau and Cucurull, Guillem and Gonz{\`a}lez, Jordi and Gonfaus, Josep M and Nasrollahi, Kamal and Moeslund, Thomas B and Roca, F Xavier},
  journal={IEEE transactions on cybernetics},
  volume={52},
  number={5},
  pages={3314--3324},
  year={2017},
  publisher={IEEE}
}

@article{au,
  title={Facial action coding system},
  author={Ekman, Paul and Friesen, Wallace V},
  journal={Environmental Psychology \& Nonverbal Behavior},
  year={1978}
}

@inproceedings{kaltwang2012continuous,
  title={Continuous pain intensity estimation from facial expressions},
  author={Kaltwang, Sebastian and Rudovic, Ognjen and Pantic, Maja},
  booktitle={International Symposium on Visual Computing},
  pages={368--377},
  year={2012},
  organization={Springer}
}

@inproceedings{yang2018lowhigh,
  title={Incorporating high-level and low-level cues for pain intensity estimation},
  author={Yang, Ruijing and Hong, Xiaopeng and Peng, Jinye and Feng, Xiaoyi and Zhao, Guoying},
  booktitle={2018 24th International Conference on Pattern Recognition (ICPR)},
  pages={3495--3500},
  year={2018},
  organization={IEEE}
}

@article{yang2021rstan,
  title={Non-contact pain recognition from video sequences with remote physiological measurements prediction},
  author={Yang, Ruijing and Guan, Ziyu and Yu, Zitong and Feng, Xiaoyi and Peng, Jinye and Zhao, Guoying},
  journal={arXiv preprint arXiv:2105.08822},
  year={2021}
}

@article{huang2021spatio,
  title={Spatio-temporal pain estimation network with measuring pseudo heart rate gain},
  author={Huang, Dong and Feng, Xiaoyi and Zhang, Haixi and Yu, Zitong and Peng, Jinye and Zhao, Guoying and Xia, Zhaoqiang},
  journal={IEEE Transactions on Multimedia},
  volume={24},
  pages={3300--3313},
  year={2021},
  publisher={IEEE}
}

@inproceedings{xu2021deep,
  title={A deep attention transformer network for pain estimation with facial expression video},
  author={Xu, Haochen and Liu, Manhua},
  booktitle={Chinese Conference on Biometric Recognition},
  pages={112--119},
  year={2021},
  organization={Springer}
}

@inproceedings{ting2021distance,
  title={Distance ordering: a deep supervised metric learning for pain intensity estimation},
  author={Ting, Jie and Yang, Yi-Cheng and Fu, Li-Chen and Tsai, Chu-Lin and Huang, Chien-Hua},
  booktitle={2021 20th IEEE International Conference on Machine Learning and Applications (ICMLA)},
  pages={1083--1088},
  year={2021},
  organization={IEEE}
}

@article{huang2022hybnet,
  title={HybNet: a hybrid network structure for pain intensity estimation},
  author={Huang, Yibo and Qing, Linbo and Xu, Shengyu and Wang, Lu and Peng, Yonghong},
  journal={The Visual Computer},
  volume={38},
  number={3},
  pages={871--882},
  year={2022},
  publisher={Springer}
}

@article{thermalvision,
  title={Thermovision: a new diagnostic method for orofacial pain?},
  author={Fricova, Jitka and Janatova, Marketa and Anders, Martin and Albrecht, Jakub and Rokyta, Richard},
  journal={Journal of Pain Research},
  pages={3195--3203},
  year={2018},
  publisher={Taylor \& Francis}
}

@article{thermalvision2,
  title={Can infrared thermography replace other methods of assessing orofacial pain intensity? Systematic review},
  author={Americano, Julia Pereira and Pires, Sofia Melo and Ferreira, Luciano Ambr{\'o}sio and Devito, Karina Lopes},
  journal={BrJP},
  volume={7},
  pages={e20240049},
  year={2024},
  publisher={SciELO Brasil}
}

@article{thermalvision3,
  title={Infrared thermography assessment of patients with temporomandibular disorders},
  author={Barbosa, JS and Amorim, AMAM and Arruda, MJALLA and Medeiros, GBS and Freitas, APLF and Vieira, LEM and Melo, DP and Bento, PM},
  journal={Dentomaxillofacial Radiology},
  volume={49},
  number={4},
  pages={20190392},
  year={2020},
  publisher={Oxford University Press}
}

@inproceedings{ytbface,
  title={Face recognition in unconstrained videos with matched background similarity},
  author={Wolf, Lior and Hassner, Tal and Maoz, Itay},
  booktitle={CVPR 2011},
  pages={529--534},
  year={2011},
  organization={IEEE}
}

@article{craig1984observer,
  title={Observer bias in judging pain in others},
  author={Craig, Kenneth D},
  journal={PAIN},
  volume={18},
  pages={S14},
  year={1984},
  publisher={LWW}
}

@article{williams2002facial,
  title={Facial expression of pain: an evolutionary account},
  author={Williams, Amanda C de C},
  journal={Behavioral and brain sciences},
  volume={25},
  number={4},
  pages={439--455},
  year={2002},
  publisher={Cambridge University Press}
}

@article{trawalter2012racial,
  title={Racial bias in perceptions of others’ pain},
  author={Trawalter, Sophie and Hoffman, Kelly M and Waytz, Adam},
  journal={PloS one},
  volume={7},
  number={11},
  pages={e48546},
  year={2012},
  publisher={Public Library of Science San Francisco, USA}
}

@article{dobscha2015short,
  title={Short-term variability in outpatient pain intensity scores in a national sample of older veterans with chronic pain},
  author={Dobscha, Steven K and Morasco, Benjamin J and Kovas, Anne E and Peters, Dawn M and Hart, Kyle and McFarland, Bentson H},
  journal={Pain Medicine},
  volume={16},
  number={5},
  pages={855--865},
  year={2015}
}

@article{mende2019perceptual,
  title={Perceptual contributions to racial bias in pain recognition.},
  author={Mende-Siedlecki, Peter and Qu-Lee, Jennie and Backer, Robert and Van Bavel, Jay J},
  journal={Journal of Experimental Psychology: General},
  volume={148},
  number={5},
  pages={863},
  year={2019},
  publisher={American Psychological Association}
}

@article{schneider2012individual,
  title={Individual differences in the day-to-day variability of pain, fatigue, and well-being in patients with rheumatic disease: associations with psychological variables},
  author={Schneider, Stefan and Junghaenel, Doerte U and Keefe, Francis J and Schwartz, Joseph E and Stone, Arthur A and Broderick, Joan E},
  journal={Pain{\textregistered}},
  volume={153},
  number={4},
  pages={813--822},
  year={2012},
  publisher={Elsevier}
}

@article{kreibig2010autonomic,
  title={Autonomic nervous system activity in emotion: A review},
  author={Kreibig, Sylvia D},
  journal={Biological psychology},
  volume={84},
  number={3},
  pages={394--421},
  year={2010},
  publisher={Elsevier}
}

@article{levenson1990voluntary,
  title={Voluntary facial action generates emotion-specific autonomic nervous system activity},
  author={Levenson, Robert W and Ekman, Paul and Friesen, Wallace V},
  journal={Psychophysiology},
  volume={27},
  number={4},
  pages={363--384},
  year={1990},
  publisher={Wiley Online Library}
}

@inproceedings{ilse2018attention,
  title={Attention-based deep multiple instance learning},
  author={Ilse, Maximilian and Tomczak, Jakub and Welling, Max},
  booktitle={International conference on machine learning},
  pages={2127--2136},
  year={2018},
  organization={PMLR}
}

@article{dietterich1997solving,
  title={Solving the multiple instance problem with axis-parallel rectangles},
  author={Dietterich, Thomas G and Lathrop, Richard H and Lozano-P{\'e}rez, Tom{\'a}s},
  journal={Artificial intelligence},
  volume={89},
  number={1-2},
  pages={31--71},
  year={1997},
  publisher={Elsevier}
}

@inproceedings{chen2020simple,
  title={A simple framework for contrastive learning of visual representations},
  author={Chen, Ting and Kornblith, Simon and Norouzi, Mohammad and Hinton, Geoffrey},
  booktitle={International conference on machine learning},
  pages={1597--1607},
  year={2020},
  organization={PmLR}
}
}

\end{document}